\documentclass{article}

\usepackage[main, final]{neurips_2026}
\setcitestyle{numbers,square}

\usepackage[utf8]{inputenc}
\usepackage[T1]{fontenc}
\usepackage[hypertexnames=false]{hyperref}
\usepackage{url}
\usepackage{booktabs}
\usepackage{float}
\usepackage{amsfonts}
\usepackage{nicefrac}
\usepackage{microtype}
\usepackage{multirow}
\usepackage[table]{xcolor}
\usepackage{tabularx}
\usepackage{enumitem}

\usepackage{amsmath}
\usepackage{amssymb}
\usepackage{amsthm}
\usepackage{mathtools}
\usepackage{bm}
\usepackage[nameinlink,noabbrev]{cleveref}
\usepackage{hyperref}
\usepackage{xurl}

\theoremstyle{plain}
\newtheorem{theorem}{Theorem}[section]

\newtheorem{lemma}[theorem]{Lemma}

\theoremstyle{definition}
\newtheorem{definition}[theorem]{Definition}

\theoremstyle{remark}

\title{Rank-Aware Differentially Private Release of Listwise Preferences for LLM Alignment}

\author{
  Junwei Chen$^{1}$,
  Manjiang Yu$^{2}$,
  Pengpeng Qiao$^{1}$,
  Yang Cao$^{1}$ \\
  $^{1}$ Institute of Science Tokyo
  \qquad
  $^{2}$ University of Queensland \\
}

\begin{document}
\raggedbottom

\maketitle

\begin{abstract}
  Listwise preference feedback offers richer supervision for large language model (LLMs) alignment than pairwise comparisons, but a full ranking also reveals more sensitive user preferences. Existing privacy-preserving alignment methods focus mainly on pairwise feedback, while listwise alignment under local differential privacy (LDP) remains unexplored. Extending LDP to this listwise domain raises a critical challenge: the high sensitivity of the full ranking space demands excessive noise, neutralizing the benefits of listwise supervision. To address this challenge, we analyze the interaction between supervision granularity and privacy noise in downstream training, and introduce a rank-aware exponential mechanism that privatizes listwise preference data into a low-sensitivity granularity suitable for downstream alignment. The mechanism leverages ranking information to sample a fixed-size binary partition, concentrating the noise near borderline items instead of perturbing all items uniformly as randomized response (RR) does. Empirical evaluations on downstream LLM alignment tasks show that our mechanism consistently outperforms existing LDP baselines at matched privacy budgets. The code is available at \url{https://github.com/tdsai-lab/RAEM}.
\end{abstract}

\section{Introduction}

\noindent
Aligning large language models (LLMs) with human preferences is a central step in post-training. Existing alignment methods often rely on pairwise feedback, such as RLHF~\citep{RLHF} and DPO~\citep{DPO}, where one response is preferred over another under the same prompt.  While effective, pairwise supervision only captures local comparisons and discards the richer relative structure among multiple candidate responses. This motivates listwise preference alignment, such as PPA~\citep{zhao-etal-2025-permutative}, LiPO~\citep{liu-etal-2025-lipo}, where several responses are evaluated jointly to provide a more informative supervision signal.

\begin{figure*}[t]
  \centering
  \includegraphics[width=0.95\textwidth]{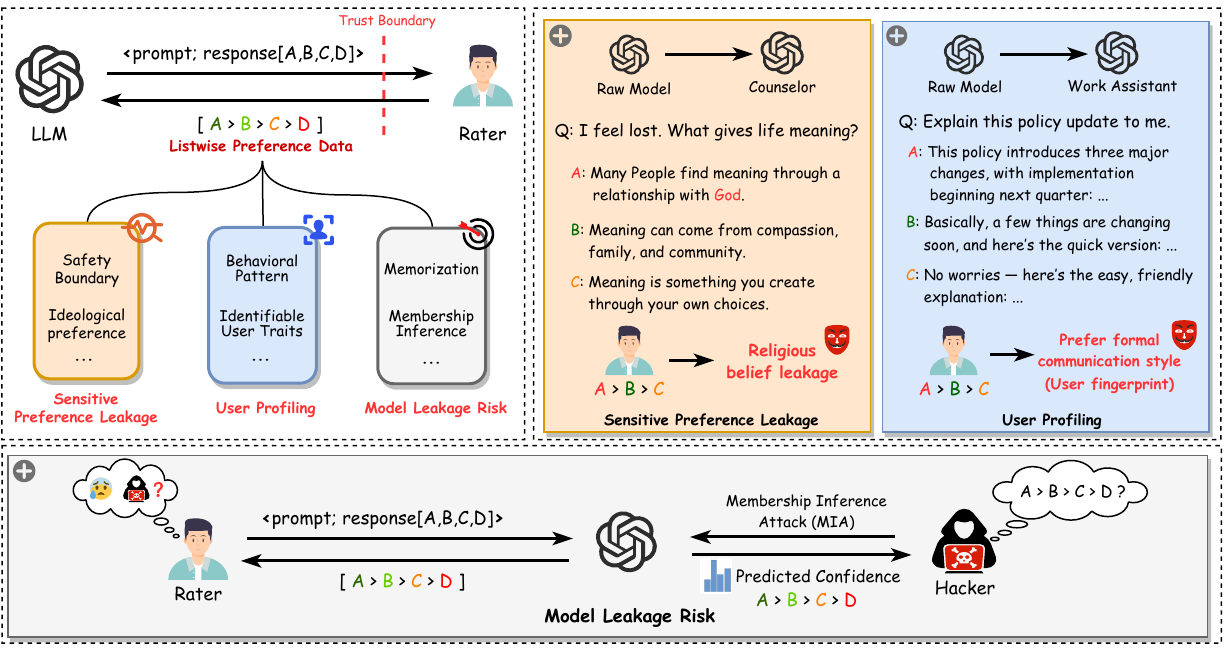}
  \caption{Threat model for privacy-preserving listwise preference alignment.}
  \label{fig:threat-model}
\end{figure*}

\noindent
Preference signals are inherently privacy-sensitive, as they may expose a user's values, beliefs, tastes, and behavioral patterns~\citep{Privacy-preserving_alignment_survey, li2026privacy}. Listwise preference data further amplifies these risks by capturing richer relative judgments over multiple candidate responses than pairwise feedback. As shown in \Cref{fig:threat-model}, if the model trainer is not fully trusted, directly releasing raw listwise feedback can cause sensitive preference leakage or user profiling. Even when the trainer is trusted, preference information may still leak through the trained model. Recent work on membership inference attacks (MIA) in alignment settings suggests that preference data can be indirectly exposed after training~\citep{feng2024exposing, liu2025grpo}. As preference alignment is increasingly used in personalized scenarios such as recommendation systems~\citep{li2025listwise, wang-etal-2025-user}, these privacy risks become more salient.

\noindent
Existing privacy-preserving alignment methods mainly focus on pairwise preference data, with LDP approaches mainly using randomized response (RR)~\cite{chaudhuri2020randomized}, where privacy is defined over binary comparisons. But listwise preference data is different: it contains scores, rankings, or group-level preferences over multiple candidates under the same prompt. This requires privacy to be formulated over the whole listwise feedback object, with sensitivity and noise calibrated to list-level changes. As a result, its sensitivity under local differential privacy can be much larger than that of a binary pairwise label. Directly privatizing the full score or ranking pattern requires substantial noise, which can destroy the ordering information that makes listwise supervision useful in the first place. Thus, a key challenge for private listwise alignment is how to inject privacy noise while preserving enough ranking information for downstream training.

\noindent
In this paper, we study listwise preference alignment under LDP. Our key idea is to use the rater's local ranking only before privatization, inside a client-side sampling mechanism, and release only a low-sensitivity binary partition to the server. This allows the mechanism to exploit ordinal information locally while exposing only the signal granularity required for downstream training after privatization. Our contributions are as follows:
\begin{itemize}[leftmargin=*,nosep]
  \item We initiate the study of private alignment under listwise feedback in the LDP setting, and introduce an item-level adjacency notion suited to ranking data.

  \item We identify release granularity as a key factor in the privacy--utility trade-off for listwise feedback, motivating a design that exploits ordinal information locally while releasing only low-sensitivity supervision for downstream training.

  \item We propose a rank-aware exponential mechanism that releases a private binary partition, assigning rank-dependent privacy noise instead of perturbing all binary labels uniformly as in RR.

  \item We prove the privacy guarantee and a sensitivity-optimality result for the proposed rank utility, and empirically show consistent gains over RR under matched privacy budgets.
\end{itemize}

\section{Related Works}

\subsection{Preference Alignment under Listwise Feedback}

Recent work has widely studied preference alignment with multiple responses for the same prompt, using scores or rankings instead of a single pairwise comparison. RRHF~\citep{yuan2023rrhf}, SLiC~\citep{zhao2023slichfsequencelikelihoodcalibration}, and PRO~\citep{song2024pro} mainly exploit ordinal rankings by encouraging the model to assign higher likelihoods to more preferred responses. LiPO-$\lambda$~\citep{liu-etal-2025-lipo} and PPA~\citep{zhao-etal-2025-permutative} use finer-grained listwise signals, such as LambdaRank-style weighting~\citep{lambdarank} and NDCG-based listwise utilities~\citep{NDCG}, to reflect ranking quality and preference strength. In contrast, KTO~\citep{KTO} only requires unary desirable or undesirable labels, making it convenient for multi-response data without constructing explicit preference pairs, but it discards ordering information within the positive and negative groups. Our work uses listwise rankings locally before privatization, while releasing only unary labels for downstream KTO training.

\subsection{Differentially Private Preference Alignment}

Differential privacy has been used to reduce leakage of individual preference during alignment~\citep{xu2025continuous}. In the central-DP setting, Yu et al.~\citep{yu2024dp} generate differentially private synthetic instructions to reduce exposure of real user feedback, while Gao et al.~\citep{gao2026differentially} propose DPPrefSyn, which privately learns a preference model from sensitive pairwise feedback and uses it with public prompts to synthesize differentially private preference data for alignment. Wu et al.~\citep{wu2023privately} apply DP mechanisms such as DP-SGD across alignment stages to mitigate training-data leakage and membership inference risks. In the local-DP setting, Teku et al.~\citep{teku2025props} propose PROPS, which protects users' preference labels under local privacy and uses progressively aligned models to generate additional feedback for later training stages. Chowdhury et al.~\citep{chowdhury2024differentially} study robust objectives for learning from randomized-response-perturbed preferences. However, these works focus on pairwise feedback and do not address locally private listwise release, where the mechanism must account for the structure and sensitivity of the entire ranking while exposing only low-sensitivity supervision for training.
%\------------------------------------------------------

\section{Preliminaries}
\label{sec:prelim}

\subsection{Exponential Mechanism}
\label{subsec:prelim-ldp-exp}

% \begin{definition}[$\varepsilon$-local differential privacy]
% \label{def:ldp}
% A randomized mechanism \(\mathcal{M}:\mathcal{X}\to\mathcal{Z}\) satisfies
% \(\varepsilon\)-local differential privacy if for all
% \(x,x'\in\mathcal{X}\) and all measurable \(\mathcal{O}\subseteq\mathcal{Z}\),
% \[
% \Pr[\mathcal{M}(x)\in\mathcal{O}]
% \le
% e^\varepsilon
% \Pr[\mathcal{M}(x')\in\mathcal{O}].
% \]
% \end{definition}

% In our setting, the private input is the clean ranking \(\phi\), and the output
% is a privatized signal released to the training server.

\begin{definition}[Exponential mechanism]
  \label{def:exponential-mechanism}
  Let \(u:\mathcal{X}\times\mathcal{Z}\to\mathbb{R}\) be a utility function with
  sensitivity
  \[
    \Delta_u
    =
    \sup_{x,x'\in\mathcal{X}}
    \sup_{z\in\mathcal{Z}}
    |u(x,z)-u(x',z)|.
  \]
  The exponential mechanism samples \(z\in\mathcal{Z}\) with probability
  \[
    \Pr[\mathcal{M}_u(x)=z]
    =
    \frac{
      \exp\!\left(\frac{\varepsilon u(x,z)}{2\Delta_u}\right)
    }{
      \sum_{z'\in\mathcal{Z}}
      \exp\!\left(\frac{\varepsilon u(x,z')}{2\Delta_u}\right)
    }.
  \]
  The exponential mechanism is known to satisfy \(\varepsilon\)- differential privacy.
\end{definition}

\subsection{Kahneman--Tversky Optimization}
\label{subsec:prelim-kto}

Kahneman--Tversky Optimization (KTO)~\citep{KTO} learns from
\emph{unary} desirability labels rather than pairwise preferences. For a
prompt--response pair \((x,y)\), let \(z\in\{0,1\}\) indicate whether \(y\) is
desirable (\(z{=}1\)) or undesirable (\(z{=}0\)). Given a policy
\(\pi_\theta\) and a fixed reference policy \(\pi_{\mathrm{ref}}\), KTO is
built on two quantities: the policy--reference log-ratio
\(r_\theta\) and a detached KL baseline \(c_\theta\),
\begin{equation}
  \label{eq:kto-ratio}
  r_\theta(x,y) \;:=\; \log\frac{\pi_\theta(y\mid x)}{\pi_{\mathrm{ref}}(y\mid x)},
  \qquad
  c_\theta \;:=\; \operatorname{sg}\!\Bigl[\,
    \mathbb{E}_{(x',y')\sim\mathcal D}\, r_\theta(x',y')
  \,\Bigr],
\end{equation}
where \(\operatorname{sg}[\cdot]\) is the stop-gradient operator. The
per-example loss interpolates between two utility terms gated by~\(z\):
\begin{equation}
  \label{eq:kto-loss}
  \ell_{\mathrm{KTO}}(\theta;x,y,z)
  \;=\;
  \underbrace{\,z\,\lambda_D\bigl[1-\sigma(\beta(r_\theta-c_\theta))\bigr]\,}_{\text{desirable}~(z=1)}
  \;+\;
  \underbrace{\,(1-z)\,\lambda_U\bigl[1-\sigma(\beta(c_\theta-r_\theta))\bigr]\,}_{\text{undesirable}~(z=0)},
\end{equation}
where \(\beta>0\) sets the log-ratio scale and \(\lambda_D,\lambda_U>0\) weight
the two regimes. Given a unary dataset
\(\mathcal D=\{(x_n,y_n,z_n)\}_{n=1}^N\), KTO minimizes the empirical risk
\begin{equation}
  \label{eq:kto-objective}
  \mathcal L_{\mathrm{KTO}}(\theta)
  \;=\;
  \frac{1}{N}\sum_{n=1}^{N} \ell_{\mathrm{KTO}}(\theta;x_n,y_n,z_n).
\end{equation}

In our setting, the clean label \(z\) is \emph{not} directly observed. It is derived from a locally privatized signal.

\section{Methodology}
\label{sec:methodology}

%------------------------------------------------------
% \subsection{Naive Score Perturbation with Laplace Noise}
% \label{subsec:naive-laplace-score-release}

% A straightforward way to privatize listwise preference feedback is to add noise directly to the raw cardinal scores. Suppose each prompt is associated with \(m\) candidate responses and a score vector
% \[
% \mathbf{s}=(s_1,\dots,s_m)\in[a,b]^m,
% \]
% where \(s_i\) denotes the preference score of response \(y_i\). The naive Laplace mechanism releases
% \[
% \widetilde{\mathbf{s}}
% =
% \mathbf{s}
% +
% \boldsymbol{\xi},
% \qquad
% \xi_i\overset{\mathrm{i.i.d.}}{\sim}
% \operatorname{Lap}\left(
% 0,
% \frac{\Delta}{\varepsilon}
% \right).
% \]

% Under item-level privacy for the whole listwise record, the score-level \(\ell_1\)-sensitivity is
% \[
% \Delta
% =
% \sup_{\mathbf{s},\mathbf{s}'\in[a,b]^m}
% \|\mathbf{s}-\mathbf{s}'\|_1
% =
% m(b-a).
% \]
% Therefore, the injected noise scale is
% \[
% \frac{\Delta}{\varepsilon}
% =
% \frac{m(b-a)}{\varepsilon}.
% \]

% By the standard Laplace mechanism, this release satisfies \(\varepsilon\)-item-level local differential privacy for the score vector.
% \[
% \widetilde{\phi}
% =
% \operatorname{Sort}(\widetilde{\mathbf{s}}),
% \]

\subsection{Item-Level Local Differential Privacy on Rankings}
\label{subsec:item-level-ldp-ranking}

Before introducing our methodology, we define local differential privacy directly over the ranking space. The atomic neighboring relation is an item-level relocation.

\begin{definition}[Item-level relocation adjacency]
  \label{def:item-relocation-adjacency}
  Two rankings \(\phi,\phi'\in S_m\) are item-level adjacent, denoted by
  \[
    \phi\sim_{\mathrm{item}}\phi',
  \]
  if \(\phi'\) can be obtained from \(\phi\) by removing one item from its current position and inserting it into another position, while preserving the relative order of all remaining items.
\end{definition}

\begin{definition}[Item-level local differential privacy on rankings]
  \label{def:item-level-ranking-ldp}
  Let \(Q:S_m\to\mathcal{O}\) be a randomized mechanism over rankings. We say that \(Q\) satisfies \(\varepsilon\)-item-level local differential privacy if for all item-level adjacent rankings \(\phi,\phi'\in S_m\) and all outputs \(o\in\mathcal{O}\),
  \[
    \frac{Q(o\mid \phi)}{Q(o\mid \phi')}
    \le
    e^\varepsilon.
  \]
\end{definition}

The guarantee in Definition~\ref{def:item-level-ranking-ldp} is a per-list item-level LDP. Intuitively, it guarantees that the exact position of any individual item in the ranking cannot be inferred by an adversary. The rank-level and user-level extension are discussed in Appendix~\ref{app:privacy-granularities}.

\subsection{Release Granularity and Privacy Noise}
\label{subsec:representation-motivation}

\noindent
In this subsection, we study the granularity of listwise preference data in
alignment scenarios and how different granularity levels are affected by
differential privacy noise, with the goal of choosing a suitable release
granularity and privacy mechanism under local privacy constraints.

\noindent
Listwise preference data can be released at several granularities: cardinal
scores provide the finest feedback, ordinal rankings keep only the relative
order, and binary top-\(B\) labels retain only whether each candidate belongs to
the preferred set. These choices preserve different amounts of preference
information and induce different local sensitivities under private release.

\begin{table}[H]
  \centering
  \small
  \setlength{\abovecaptionskip}{4pt}
  \setlength{\belowcaptionskip}{-2pt}
  \setlength{\tabcolsep}{4pt}
  \renewcommand{\arraystretch}{1.00}
  \begin{tabularx}{0.98\linewidth}{
      @{\hspace{2pt}}
      >{\raggedright\arraybackslash}p{0.30\linewidth}
      >{\centering\arraybackslash}p{0.25\linewidth}
      >{\centering\arraybackslash}p{0.19\linewidth}
      >{\centering\arraybackslash}p{0.20\linewidth}
      @{\hspace{2pt}}
    }
    \toprule
    \textbf{Supervision granularity}
    & \textbf{Representative method}
    & \textbf{Sensitivity}
    & \textbf{Noise scale} \\
    \midrule
    Cardinal score vector
    & PPA~\citep{zhao-etal-2025-permutative}
    & \(m^2\)
    & \(O(m^2/\varepsilon)\) \\
    Rank vector
    & SLiC~\citep{zhao2023slichfsequencelikelihoodcalibration}
    & \(2(m-1)\)
    & \(O(m/\varepsilon)\) \\
    Binary labels
    & KTO~\citep{KTO}
    & \(2\)
    & \(O(1/\varepsilon)\) \\
    \bottomrule
  \end{tabularx}
  \caption{Data release granularity, representative alignment methods, and noise scale. Sensitivity bounds and noise scale are proved in Appendix~\ref{app:sensitivity-calculation}.}
  \label{tab:representation-granularity}
  \vspace{-6pt} % reduce gap between table float and following text
\end{table}

\noindent
\Cref{tab:representation-granularity} shows that finer-grained releases incur
higher privacy cost. While cardinal scores and full rankings preserve more
pre-privatization information, their larger sensitivities require stronger
perturbation, especially for large \(m\) or small \(\varepsilon\). This affects
downstream training: PPA and SLiC rely on reliable relative preferences across the candidate list, whereas KTO only needs unary desirable/undesirable labels. Thus, DP noise in fine-grained releases can more severely degrade supervision quality and optimization stability.

\noindent
However, simply operating on coarse-grained information also risks losing high-quality preference signals: binary top-\(B\) labels collapse the ordering within positive and negative groups, making it difficult to distinguish the best responses from merely acceptable ones. This motivates our design: we use local rankings only for client-side sampling and release a low-sensitivity binary top-\(B\) partition, providing the server with unary labels while retaining ordinal information before privatization.

\subsection{Rank-Aware Exponential Mechanism for Binary Release}

Motivated by Section~\ref{subsec:representation-motivation}, we use the exponential mechanism in Definition~\ref{def:exponential-mechanism} to privately sample low-sensitivity binary top-\(B\) partitions while favoring outputs that preserve the local ranking.

\paragraph{Problem Setup}
\noindent For each prompt \(x\), the client observes candidate responses
\(Y(x)=\{y_1,\ldots,y_m\}\) and local supervision scores
\(\mathbf{s}=(s_1,\ldots,s_m)\). The scores induce a clean ranking
\(\phi=\operatorname{Sort}(\mathbf{s})\), where \(r_i(\phi)\) denotes the rank of
response \(y_i\), with smaller ranks indicating stronger preference.

The client releases a fixed-size binary vector from
\[
  \mathcal Z_B:=\{z\in\{0,1\}^m:\|z\|_0=B\}.
\]
Given \(\phi\), the release mechanism samples
\[
  Z\mid\phi\sim Q(\cdot\mid\phi),
  \qquad Z\in\mathcal Z_B,
\]
and outputs its realization \(\widetilde z\). The server observes only
\((x,Y(x),\widetilde z)\), where \(\widetilde z_i=1\) marks \(y_i\) as desirable
and \(\widetilde z_i=0\) marks it as undesirable.

% The local scores
% \(\mathbf{s}\) and the clean ranking \(\phi\) are not released.

\paragraph{Rank-aware utility.}
We instantiate the release utility with a linear boundary score. Given a clean
ranking \(\phi\), let \(r_i(\phi)\) be the rank of response \(y_i\), with smaller
rank indicating stronger preference. The linear score assigns each item a
signed distance to the top-\(B\) boundary:

\begin{equation}
  \label{eq:rank-utility}
  w_i(\phi)=B+\frac12-r_i(\phi),
  \qquad
  U(\phi,z)
  =
  \sum_{i=1}^m z_i w_i(\phi)
  =
  \sum_{i=1}^m z_i\left(B+\frac12-r_i(\phi)\right).
\end{equation}

Thus, items above the boundary receive positive weights and items below it receive negative weights.
Maximizing \(U(\phi,z)\) favors partitions that select higher-ranked responses while penalizing selections from lower-ranked positions.
Under item-level relocation adjacency, this utility has sensitivity

\[
  \Delta
  :=
  \max_{\phi\sim\phi'}
  \max_{z\in\mathcal{Z}_B}
  |U(\phi,z)-U(\phi',z)|
  =
  \max\{B,m-B\},
\]

\noindent
This sensitivity is proved in \Cref{lem:rank-utility-sensitivity}. Section~\ref{subsec:optimal-extreme-reliability} shows that this linear boundary score is sensitivity-optimal among monotone rank scores under a normalized rank-span criterion.

% \noindent \paragraph{Private release.}
% The client samples a binary partition by exponential mechanism
% \[
%   Q(z\mid\phi)
%   =
%   \frac{
%     \exp\!\left(\eta U(\phi,z)\right)
%   }{
%     \sum_{z'\in\mathcal{Z}_B}
%     \exp\!\left(\eta U(\phi,z')\right)
%   },
%   \qquad
%   \eta
%   =
%   \frac{\varepsilon}{\Delta}
%   =
%   \frac{\varepsilon}{\max\{B,m-B\}},
% \]
% and releases the sampled \(\tilde z\sim Q(\cdot\mid\phi)\) as the private
% top-\(B\) selection. By Definition~\ref{def:exponential-mechanism}, this is
% the standard exponential mechanism with utility \(U\) and sensitivity
% \(\Delta\), and therefore satisfies \(\varepsilon\)-local differential
% privacy under the item-level ranking adjacency.

\noindent \paragraph{Private release.}
The client samples a binary partition by exponential mechanism
\[
  Q(z\mid\phi)
  =
  \frac{
    \exp\!\left(\eta U(\phi,z)\right)
  }{
    \sum_{z'\in\mathcal{Z}_B}
    \exp\!\left(\eta U(\phi,z')\right)
  },
  \qquad
  \eta
  =
  \frac{\varepsilon}{\Delta}
  =
  \frac{\varepsilon}{\max\{B,m-B\}},
\]
and releases the sampled \(\tilde z\sim Q(\cdot\mid\phi)\) as the private top-\(B\) selection. By Theorem~\ref{thm:rank-aware-release-privacy} in Appendix~\ref{app:rank-aware-release-privacy}, this exponential mechanismwith utility \(U\) and sensitivity \(\Delta\) satisfies \(\varepsilon\)-local differential privacy under the item-level ranking adjacency.

\paragraph{Server-side training.}
\noindent After release, the server observes only \((x,Y(x),\widetilde z)\) and expands
each privatized list into unary preference examples
\[
  \mathcal D_{\mathrm{priv}}(x)
  =
  \{(x,y_i,\widetilde z_i)\}_{i=1}^m,
  \qquad
  \widetilde z_i=1 \text{ desirable},\quad
  \widetilde z_i=0 \text{ undesirable}.
\]
\noindent The policy is then optimized with the KTO objective in Section~\ref{subsec:prelim-kto} using \(\mathcal D_{\mathrm{priv}}\). The server never observes the local scores \(\mathbf{s}\), the clean ranking \(\phi\), or the clean top-\(B\) labels \(z(\phi)\).

% \Cref{fig:raem-pipeline} illustrates the pipeline of our method.

% \begin{figure}[t]
%   \centering
%   \includegraphics[width=0.85\linewidth]{images/pipeline.pdf}
%   \caption{Overview of the proposed rank-aware exponential mechanism. The
%     client first ranks candidate responses locally, then uses a differentially
%     private sampler to release a fixed-size binary partition. Only the privatized
%   positive/negative labels cross the trust boundary to the server.}
%   \label{fig:raem-pipeline}
% \end{figure}

\section{Flip Rate Analysis}
\label{sec:flip-rate-design}

\noindent
The flip rate directly affects downstream KTO training because the released binary partition is used as unary supervision. A flipped label turns a positive KTO update into a negative one, or vice versa, thereby reversing the training direction of that sample. Hence, the location of privacy-induced flips matters: top-ranked positives and bottom-ranked negatives provide the most reliable
unary labels, while samples near the top-\(B\) boundary are more ambiguous. We
therefore prefer a release mechanism with low flip rates at extreme ranks and
higher flip rates near the boundary.

\subsection{Per-Rank Flip Rates of Rank-Aware Exponential Mechanism}
\label{subsec:flip-rate-formula}

\noindent Let
\(q_i(\phi):=\Pr[\tilde z_i=1\mid\phi]\) denote the marginal probability that
\(y_i\) is released as positive. Under the rank-aware exponential mechanism,
\[
  q_i(\phi)
  =
  \frac{
    e^{\eta w_i(\phi)}\,\sigma_{B-1}\!\bigl(e^{\eta w_{-i}(\phi)}\bigr)
  }{
    \sigma_B\!\bigl(e^{\eta w(\phi)}\bigr)
  },
  \qquad
  \rho_i(\phi)
  :=
  \Pr[\tilde z_i\ne z_i(\phi)\mid\phi]
  =
  \begin{cases}
    1-q_i(\phi), & r_i(\phi)\le B,\\
    q_i(\phi),   & r_i(\phi)>B,
  \end{cases}
\]
where \(\sigma_k\) is the \(k\)-th elementary symmetric polynomial,
\(e^{\eta w(\phi)}\) is applied coordinatewise, and \(w_{-i}(\phi)\) drops coordinate \(i\). The flip rate is rank-dependent: extreme ranks remain stable while flips concentrate near the top-\(B\) boundary, with derivation and monotonicity deferred to Appendix~\ref{app:flip-rate-derivation}.

\subsection{Sensitivity-Optimality of the Linear Score}
\label{subsec:optimal-extreme-reliability}

\noindent
We evaluate a monotone rank score \(w_1\ge\cdots\ge w_m\) by its
\emph{calibrated rank span} \(S_w/\Delta_w\), where \(S_w=w_1-w_m\) and
\(\Delta_w\) is the item-level relocation sensitivity of the induced
rank-based utility \(U_w\). Since the exponential mechanism sets
\(\eta=\varepsilon/\Delta_w\), the top-to-bottom logit gap is
\[
  \eta S_w=\varepsilon\cdot\frac{S_w}{\Delta_w},
\]
\noindent so maximizing \(S_w/\Delta_w\) maximizes extreme-rank separation at
any fixed privacy budget.

\begin{theorem}[Optimal calibrated rank span]
  \label{thm:linear-extreme-reliability}
  For any monotone rank score \(w_1\ge\cdots\ge w_m\),
  \[
    \frac{S_w}{\Delta_w}\;\le\;\frac{m-1}{\max\{B,m-B\}},
  \]
  with equality iff \(w_s=\alpha-\beta s\) for some \(\beta>0\).
\end{theorem}

\noindent Thus the bound is tight only for linear scores, which therefore yield the most stable extreme releases under item-level relocation adjacency. The proof is deferred to Appendix~\ref{app:proof-optimal-score-function}.

\subsection{Comparison with RR}
\label{subsec:rr-comparison}

\noindent We compare the rank-aware exponential mechanism with RR, the canonical LDP baseline for binary labels. RR privatizes the clean top-\(B\) label vector
\[
  z_i(\phi)=\mathbf{1}[r_i(\phi)\le B],
  \qquad
  z(\phi)=(z_1(\phi),\ldots,z_m(\phi)).
\]
RR is originally designed for privatizing single binary
labels. In our setting, however, each prompt produces a vector of binary labels over multiple candidate responses. Therefore, we calibrate RR to the sensitivity of the full clean top-\(B\) vector. Under item-level relocation adjacency in Definition~\ref{def:item-relocation-adjacency}, this vector has Hamming sensitivity
\[
  \Delta_H
  :=
  \sup_{\phi\sim_{\mathrm{item}}\phi'}
  \|z(\phi)-z(\phi')\|_0
  =
  2.
\]
\noindent Thus, applying coordinate-wise RR with per-coordinate privacy parameter
\(\varepsilon_0=\varepsilon/\Delta_H=\varepsilon/2\) yields
\(\varepsilon\)-LDP for the full released vector. The calibration proof is deferred to Appendix~\ref{app:rr-calibration}.

\noindent Let \(\tilde z_{\mathsf{RR}}\) denote the RR release. By definition, each coordinate flips independently with probability
\[
  \Pr[(\tilde z_{\mathsf{RR}})_i = z_i(\phi)]
  = \frac{e^{\varepsilon/2}}{1+e^{\varepsilon/2}},
  \qquad
  \rho_i^{\mathsf{RR}}(\phi)
  := \Pr[(\tilde z_{\mathsf{RR}})_i \ne z_i(\phi)]
  = \frac{1}{1+e^{\varepsilon/2}}.
\]

\begin{figure}[t]
  \centering
  \includegraphics[width=0.8\linewidth]{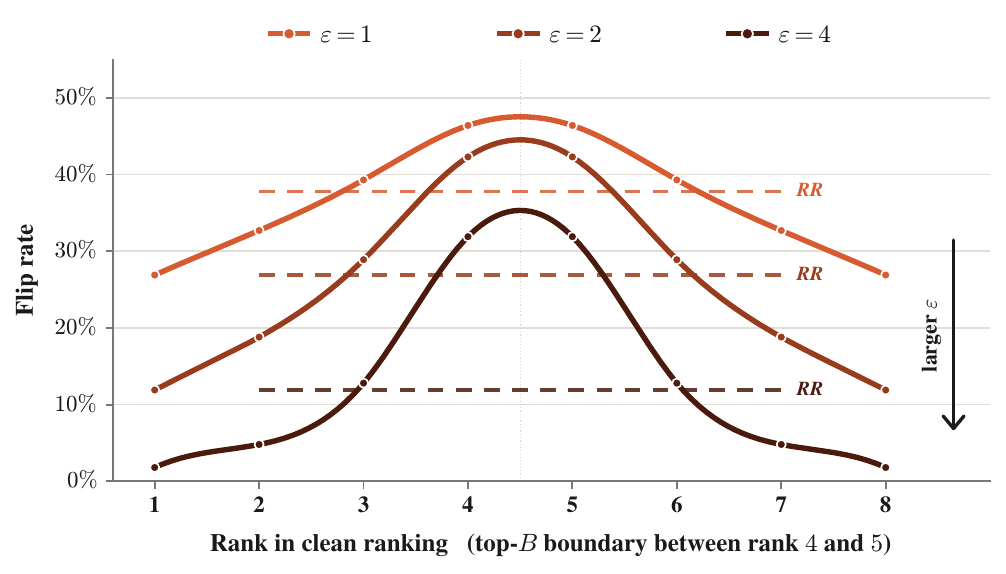}
  \caption{
    Simulation of per-rank flip rates for our method (solid curves) and RR
    (dashed lines) with \(m=8\) and \(B=4\), for different values of
    \(\varepsilon\).
  }
  \label{fig:flip-rate-profile}
\end{figure}

\noindent Thus, RR has a rank-independent flip rate, perturbing all coordinates equally. In contrast, our method induces rank-dependent noise, preserving high-quality ranks while concentrating flips near the top-\(B\) boundary. \Cref{fig:flip-rate-profile} further illustrates how privacy calibration changes the noise profile.

\section{Experiments}
\label{sec:experiments}

\subsection{Setting}

\paragraph{Dataset.}

\noindent
We use \textit{ListUltraFeedback}, an offline listwise preference dataset constructed in PPA~\citep{zhao-etal-2025-permutative}.
It is built from \textit{UltraFeedback}~\citep{cui2023ultrafeedback} by combining \textit{UltraFeedback} responses with additional SimPO-Llama3-8B\footnote{\url{https://huggingface.co/princeton-nlp/Llama-3-Instruct-8B-SimPO}} responses and scoring them with ArmoRM\footnote{\url{https://huggingface.co/RLHFlow/ArmoRM-Llama3-8B-v0.1}} to obtain listwise preferences.
The dataset contains 59.9K training examples and 1,968 held-out examples.

\paragraph{Baselines.}

\noindent
We compare with non-private alignment methods as utility references, including PPA, SLiC~\citep{zhao2023slichfsequencelikelihoodcalibration}, and KTO~\citep{KTO}. These methods use clean preference supervision without privacy constraints and therefore quantify the utility loss introduced by private release. As the main private baseline, we use RR followed by KTO. We select PPA over LiPO~\citep{liu-etal-2025-lipo} based on its stronger performance in the PPA paper's evaluation, and omit PROPS~\citep{teku2025props} because its comparison-level privacy unit, binary-comparison supervision, and additional training stages prevent a direct comparison under our ranking-level adjacency and unary-label KTO setting.

\paragraph{Training Details.}

\noindent
We use Qwen2-0.5B~\citep{yang2024qwen2} and Mistral-7B~\citep{jiang2023mistral7b} as foundation models. For each model, we first perform supervised fine-tuning on UltraChat-200K~\citep{ding2023enhancing} to obtain the SFT checkpoint, and then initialize all alignment methods from this same checkpoint on \textit{ListUltraFeedback}. For both Qwen2-0.5B and Mistral-7B, we train all alignment methods using the AdamW optimizer~\citep{loshchilov2017decoupled} on 4 NVIDIA H100 80GB GPUs. Further training details are provided in Appendix~\ref{app:training-details}.

\paragraph{Evaluation.}

\noindent
We report win rates of aligned models against the corresponding SFT model.
Since ArmoRM is used to construct \textit{ListUltraFeedback}, evaluation by only ArmoRM may overestimate performance due to reward-model coupling. To mitigate this, we use two complementary evaluation pipelines: an independent
pair-preference reward model and general conversational benchmarks. In the
proxy-model pipeline, we compute win rates on the held-out
\textit{ListUltraFeedback} split using ArmoRM as a scoring reward model and the RLHFlow pair-preference reward model\footnote{\url{https://huggingface.co/RLHFlow/pair-preference-model-LLaMA3-8B}}
as a pairwise judge. In the general-benchmark pipeline, we evaluate
conversational quality on AlpacaEval~\citep{li2023alpacaeval} and
MT-Bench~\citep{zheng2023judging}, using GPT-4 Turbo~\citep{achiam2023gpt} as the judge following the original setup. To reduce A/B order bias, we randomly shuffle the two candidate responses before judging and map the decision back to the original model identity. Further evaluation details are provided in Appendix~\ref{app:evaluation-details}.
%=====================================================================

\subsection{Main results}

\noindent \Cref{tab:main_results} compares our method with non-private alignment references and the private RR baseline. Our method achieves the best average performance among private methods for every tested privacy budget and both backbone models. The improvement is more pronounced for Qwen2-0.5B, while on Mistral-7B the same trend holds despite the stronger SFT and non-private alignment baselines.

\noindent
Both our method and RR train KTO on noisy unary labels, but RR assigns a rank-independent flip probability, whereas our mechanism uses the local ranking to concentrate noise near the top-\(B\) boundary and preserve extreme ranks with lower flip rates. The consistent gains show that, for private KTO training, where the noise is placed is important under the same privacy budget. Allocating more flips to boundary-near examples is more effective than perturbing all binary labels uniformly. Additional AlpacaEval results are reported in Appendix~\ref{app:supplementary-results}.

\begin{table}[!t]
  \centering
  \setlength{\abovecaptionskip}{6pt}
  \small
  \setlength{\tabcolsep}{8pt}
  \renewcommand{\arraystretch}{0.85}
  \definecolor{qwenbg}{RGB}{224,226,224}     % graphite tint
  \definecolor{mistralbg}{RGB}{234,228,218}  % sand
  \definecolor{ourbg}{RGB}{236,236,236}
  \begin{tabular}{l c c c c c}
    \toprule
    \multirow{2}{*}{\textbf{Method}} &
    \multicolumn{2}{c}{\textbf{Proxy Model}} &
    \multicolumn{2}{c}{\textbf{General Benchmark}} &
    \multirow{2}{*}{\textbf{Avg.}} \\
    \cmidrule(lr){2-3}\cmidrule(lr){4-5}
    & Pair-Preference & Scoring & AlpacaEval & MT-Bench & \\
    \midrule
    % ===== Qwen2.0-0.5B =====
    \rowcolor{qwenbg}\multicolumn{6}{c}{\textbf{Qwen2-0.5B}}\\
    \midrule
    \multicolumn{6}{@{}l}{\textbf{\textit{Non-private}} ($\varepsilon{=}\infty$)}\\[2pt]
    \quad PPA  & 66.3 & 61.3 & 61.6 & 56.4 & 61.4 \\
    \quad SLiC & 65.7 & 60.7 & 61.0 & 55.8 & 60.8 \\
    \quad KTO  & 67.8 & 55.9 & 59.2 & 60.1 & 60.8 \\
    \multicolumn{6}{@{}l}{\textbf{\textit{Private}}, $\varepsilon=0.5$}\\[2pt]
    \quad RR$+$KTO   & 52.9 & 38.6 & 48.0 & 52.5 & 48.0 \\
    \rowcolor{ourbg}\quad RAEM$+$KTO & 59.2 & 43.6 & 51.8 & 50.1 & \textbf{51.2} \\
    \multicolumn{6}{@{}l}{\textbf{\textit{Private}}, $\varepsilon=1$}\\[2pt]
    \quad RR$+$KTO   & 57.9 & 43.7 & 50.7 & 52.0 & 51.1 \\
    \rowcolor{ourbg}\quad RAEM$+$KTO & 61.2 & 49.1 & 50.6 & 50.6 & \textbf{52.9} \\
    \multicolumn{6}{@{}l}{\textbf{\textit{Private}}, $\varepsilon=3$}\\[2pt]
    \quad RR$+$KTO   & 64.4 & 51.5 & 56.5 & 56.0 & 57.1 \\
    \rowcolor{ourbg}\quad RAEM$+$KTO & 66.4 & 54.6 & 57.8 & 59.7 & \textbf{59.6} \\
    \midrule
    % ===== Mistral-7B =====
    \rowcolor{mistralbg}\multicolumn{6}{c}{\textbf{Mistral-7B}}\\
    \midrule
    \multicolumn{6}{@{}l}{\textbf{\textit{Non-private}} ($\varepsilon{=}\infty$)}\\[2pt]
    \quad PPA  & 91.5 & 93.1 & 90.2 & 73.8 & 87.2 \\
    \quad SLiC & 92.4 & 92.9 & 89.8 & 72.8 & 87.0 \\
    \quad KTO  & 92.5 & 92.7 & 90.8 & 72.3 & 87.1 \\
    \multicolumn{6}{@{}l}{\textbf{\textit{Private}}, $\varepsilon=0.5$}\\[2pt]
    \quad RR$+$KTO   & 88.8 & 88.4 & 85.9 & 73.0 & 84.0 \\
    \rowcolor{ourbg}\quad RAEM$+$KTO & 90.4 & 90.6 & 87.0 & 72.8 & \textbf{85.2} \\
    \multicolumn{6}{@{}l}{\textbf{\textit{Private}}, $\varepsilon=1$}\\[2pt]
    \quad RR$+$KTO   & 89.7 & 90.9 & 86.4 & 71.8 & 84.7 \\
    \rowcolor{ourbg}\quad RAEM$+$KTO & 90.5 & 91.0 & 87.5 & 72.1 & \textbf{85.3} \\
    \multicolumn{6}{@{}l}{\textbf{\textit{Private}}, $\varepsilon=3$}\\[2pt]
    \quad RR$+$KTO   & 91.4 & 92.1 & 87.4 & 72.1 & 85.8 \\
    \rowcolor{ourbg}\quad RAEM$+$KTO & 91.5 & 92.4 & 90.4 & 72.4 & \textbf{86.7} \\
    \bottomrule
  \end{tabular}
  \caption{\textbf{Main results} with listwise preference alignment (\(m{=}8\), \(B{=}4\)).
  \textbf{Avg.} denotes the simple mean of the four reported metrics. Shaded rows (RAEM+KTO) denote our method.}
  \label{tab:main_results}
\end{table}
% =====================================================================

\subsection{Error-to-correct Flip Rate}
\label{subsec:flip-rate}

\begin{figure*}[!t]
  \centering
  \begin{minipage}[!t]{0.49\textwidth}
    \centering
    \includegraphics[width=\linewidth]{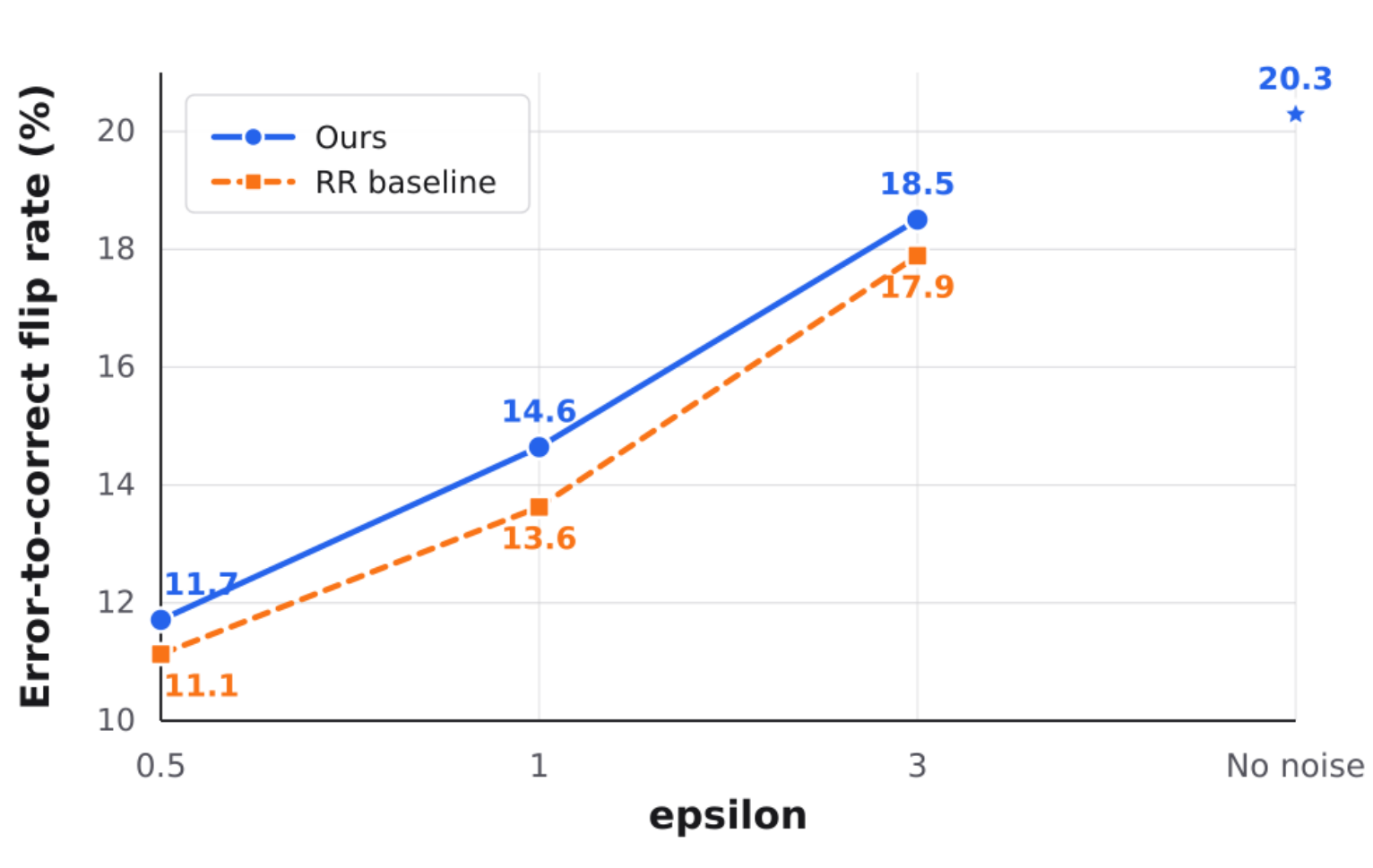}

    {\small \textbf{(a)} Qwen2-0.5B}
  \end{minipage}\hfill
  \begin{minipage}[!t]{0.49\textwidth}
    \centering
    \includegraphics[width=\linewidth]{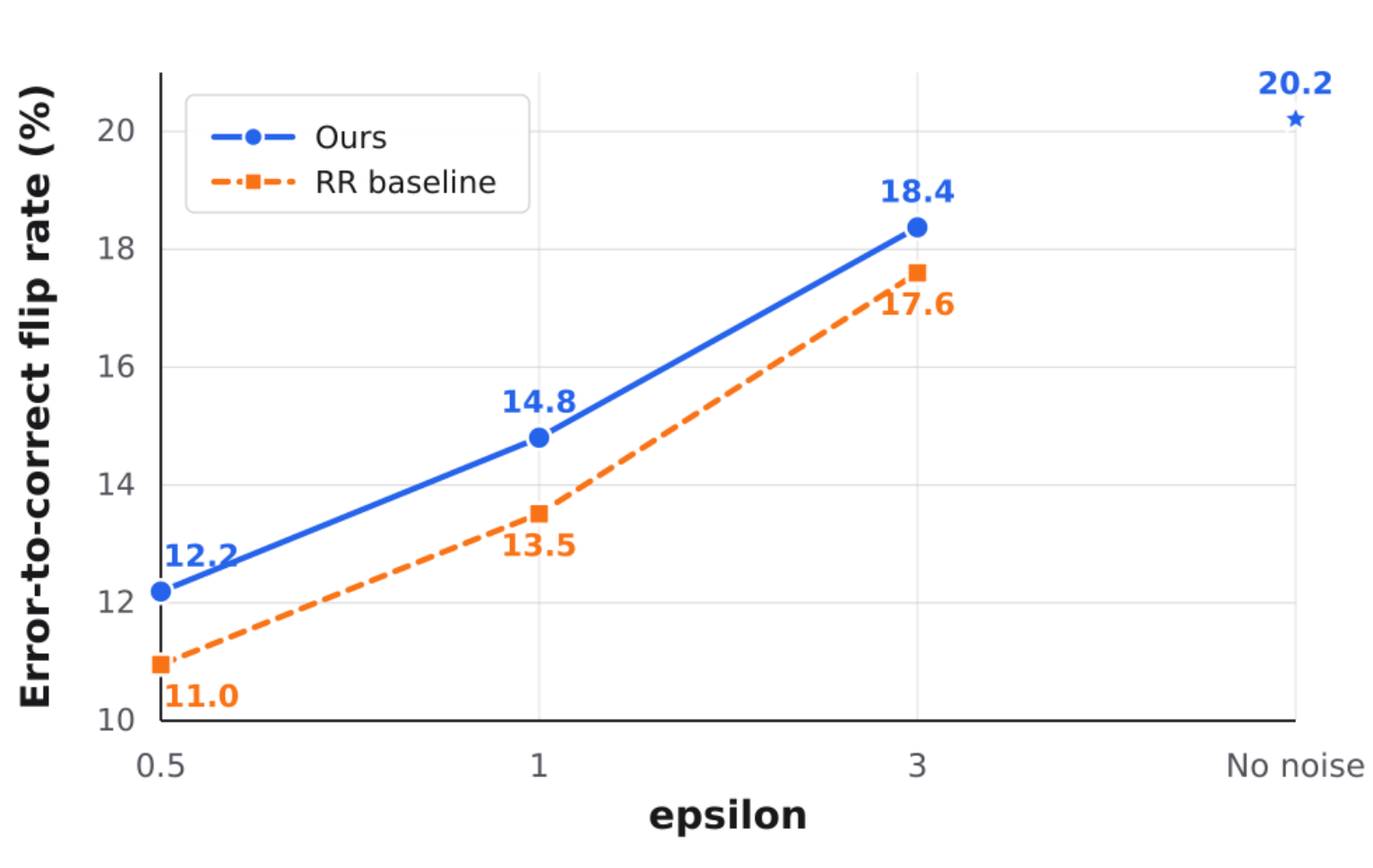}

    {\small \textbf{(b)} Mistral-7B}
  \end{minipage}
  \caption{
    Error-to-correct flip rate (E2C) of our method and the RR baseline on both models. The starred point marks the no-noise reference.
  }
  \label{fig:flip-rate-experiments}
\end{figure*}

\noindent
We evaluate error-to-correct flip rate (E2C), the fraction of SFT-induced wrong pairwise
preferences corrected after alignment. Pairwise preferences are induced from
response log-probabilities:
\[
  y_i\succ_\theta y_j
  \quad\Longleftrightarrow\quad
  \log p_\theta(y_i\mid x) > \log p_\theta(y_j\mid x).
\]
Higher E2C indicates more corrected initial preference errors. \noindent
\Cref{fig:flip-rate-experiments} shows that our method consistently achieves
higher E2C than RR and stays closer to the no-noise reference, suggesting that the rank-aware exponential mechanism preserves stronger correction signals.

\subsection{Effect of List Size}
\label{subsec:effect-of-list-size}

\noindent We analyze whether full-list private release preserves high-quality training signals better than selecting a small candidate subset before release, using Qwen2-0.5B as the backbone model.

\noindent \Cref{fig:candidate-selection} illustrates the role of list size under private release. In the ideal oracle setting, the learner keeps the clean top-ranked and bottom-ranked responses, which provide the strongest KTO supervision (\Cref{fig:candidate-selection}(a)). Under this selection rule, increasing the candidate list size is not always beneficial: additional boundary-near, low-confidence responses introduce more released labels and hence more privacy
noise, without necessarily improving the best positive or worst negative examples. However, this oracle selection is not a valid private procedure, because identifying the clean extremes already reveals information from the user's ranking.

\noindent In a strict privacy-preserving setting, a valid small-subset method must select candidates before observing the clean ranking. Without prior quality information, this reduces to random sampling, which may miss high-quality extremes. As shown by the random-subset curve in \Cref{fig:candidate-selection}(a), larger subsets increase the chance of including strong positives and negatives, but the sampling risk remains. Our method avoids this risk by applying private release to the full list.

\noindent Larger listwise preference sets are more likely to contain high-quality extreme responses. Meanwhile, \Cref{fig:candidate-selection}(b) shows that our method keeps the flip rates of the top and bottom ranks stable as \(m\) increases, with additional privacy noise absorbed by boundary-near responses. Thus, in the private setting, increasing \(m\) improves the available listwise supervision while our mechanism prevents the most valuable examples from becoming less reliable under LDP.

\noindent Additional experiments on DPO compatibility, training-seed variability,
and mechanism ablations are reported in
Appendix~\ref{app:additional-experiments}.

% 放在导言区
\newlength{\panelgraphicheight}

% 正文
\begin{figure*}[!t]
  \centering
  % 右图以 0.48\textwidth 显示时，高度约为 0.375\textwidth
  \setlength{\panelgraphicheight}{0.38\textwidth}

  \begin{minipage}[t]{0.48\textwidth}
    \vspace{0pt}
    \centering
    \begin{minipage}[t][\panelgraphicheight][t]{\linewidth}
      \vspace{0pt}
      \centering
      \includegraphics[width=\linewidth]{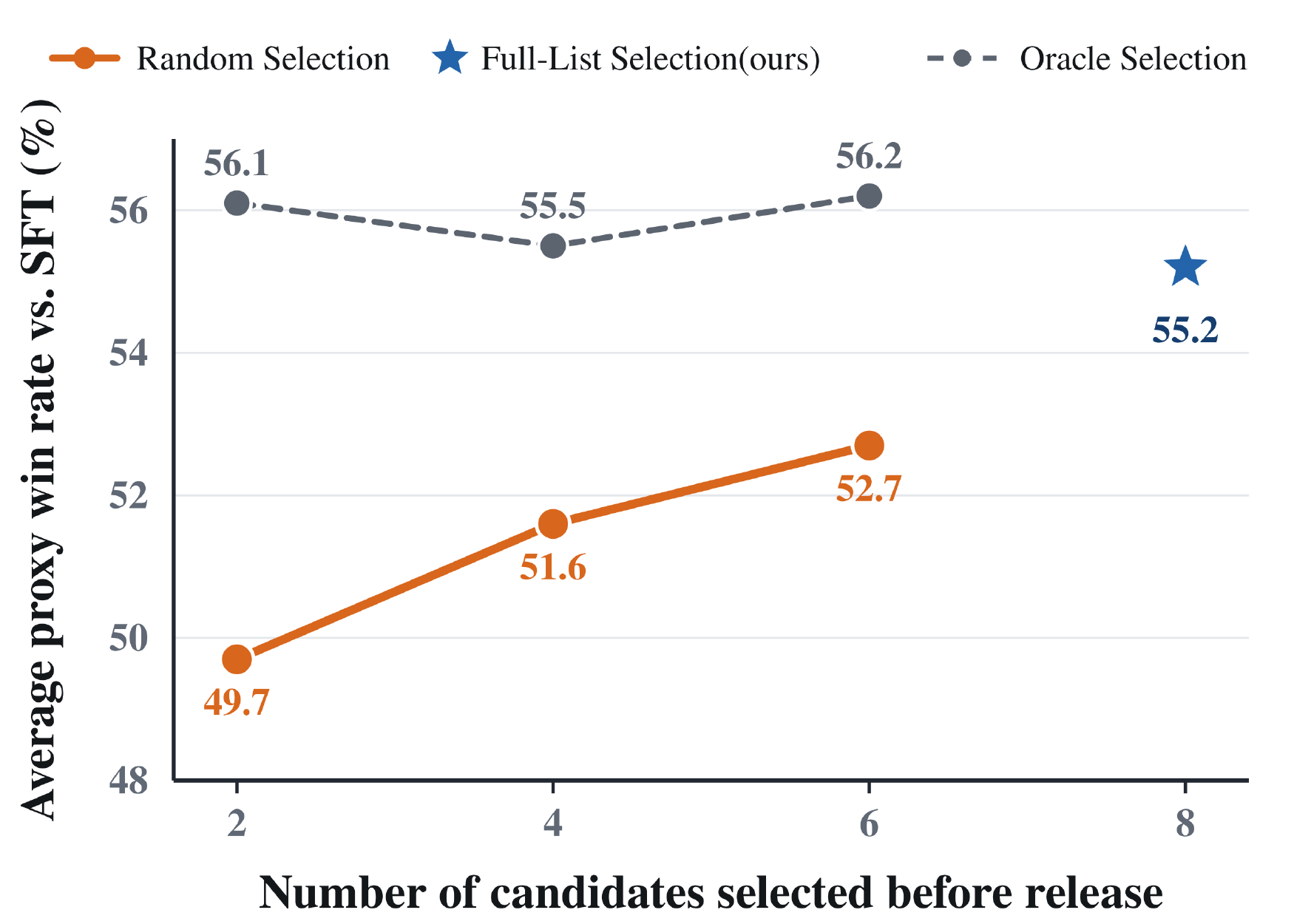}
    \end{minipage}

    \smallskip
    {\small \textbf{(a)} Candidate selection before private release}
  \end{minipage}
  \hfill
  \begin{minipage}[t]{0.48\textwidth}
    \vspace{0pt}
    \centering
    \begin{minipage}[t][\panelgraphicheight][t]{\linewidth}
      \vspace{0pt}
      \centering
      \includegraphics[width=\linewidth]{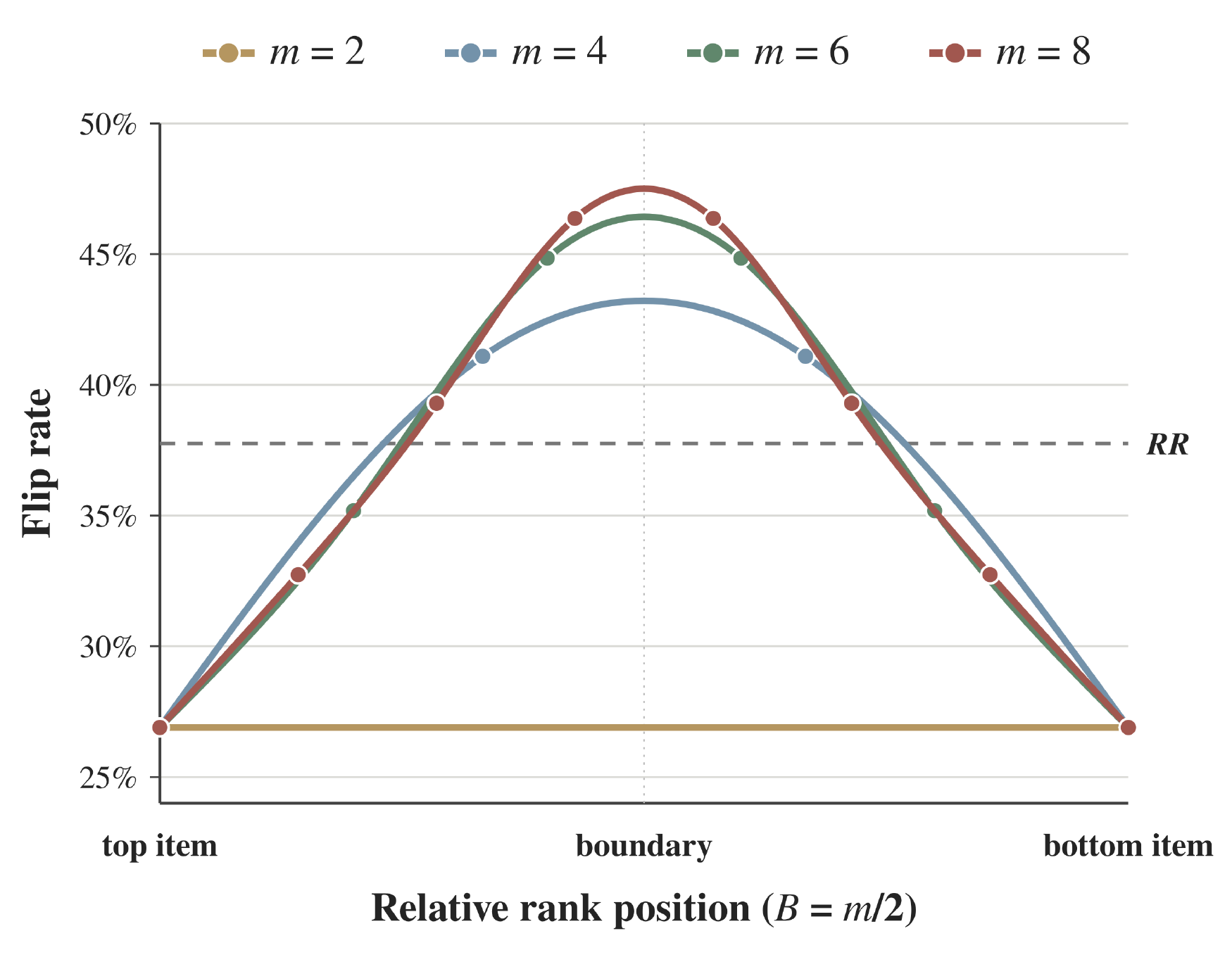}
    \end{minipage}

    \smallskip
    {\small \textbf{(b)} Flip-rate profile as \(m\) varies}
  \end{minipage}

  \caption{
    Candidate selection and flip-rate allocation at \(\varepsilon=1\).
    (a) Average proxy win rate against the corresponding SFT model after applying
    our rank-aware private release and KTO training to Qwen2-0.5B under different
    candidate-selection rules. Oracle top-bottom selection retains the clean best
    and worst responses and randomly fills the remaining slots. Random subset
    selection samples all \(k\) candidates before release, whereas our default
    method retains all candidates.
    (b) Flip-rate profiles of the rank-aware exponential mechanism concentrate
    near the top-\(B\) boundary as \(m\) increases.
  }
  \label{fig:candidate-selection}
\end{figure*}

\section{Limitation}

\noindent
Our analysis focuses on item-level local differential privacy in Definition~\ref{def:item-level-ranking-ldp}. Under item-level
relocation adjacency, the guarantee protects the position of any individual response: two rankings that differ only by relocating one item cannot be easily distinguished from the released signal. This captures the main leakage risk in a listwise feedback setting. However, optimizing for these stricter notions is outside the scope of this work. Appendix~\ref{app:privacy-granularities}
defines stricter rank-level and user-level variants, which protect an entire ranking or all rankings from the same user. Optimizing for these stronger notions
is left for future work, since direct extensions require calibration to linear composition over multiple released lists, causing a large privacy-budget cost, especially at the user level.

\section{Conclusion}

\noindent
We studied listwise preference alignment under local differential privacy. We showed that releasing scores or full rankings incurs high sensitivity, and proposed a rank-aware exponential mechanism that uses rankings locally while releasing only binary top-\(B\) labels. Theoretical and empirical results show that our method better preserves preference signals than RR under matched privacy budgets, highlighting that private alignment depends not only on noise scale but also on where errors occur.

\begin{ack}

This work was supported by JST PRESTO (Grant No. JPMJPR23P5), JST CREST (Grant No. JPMJCR21M2), JSPS KAKENHI (Grant No. 26K21227) and JST NEXUS (Grant No. JPMJNX25C4).

\end{ack}

% \clearpage
{\small
  \bibliographystyle{plainnat}
  \bibliography{references}

@inproceedings{RLHF,
 author = {Ouyang, Long and Wu, Jeffrey and Jiang, Xu and Almeida, Diogo and Wainwright, Carroll and Mishkin, Pamela and Zhang, Chong and Agarwal, Sandhini and Slama, Katarina and Ray, Alex and Schulman, John and Hilton, Jacob and Kelton, Fraser and Miller, Luke and Simens, Maddie and Askell, Amanda and Welinder, Peter and Christiano, Paul F and Leike, Jan and Lowe, Ryan},
 booktitle = {Advances in Neural Information Processing Systems},
 editor = {S. Koyejo and S. Mohamed and A. Agarwal and D. Belgrave and K. Cho and A. Oh},
 pages = {27730--27744},
 publisher = {Curran Associates, Inc.},
 title = {Training language models to follow instructions with human feedback},
 url = {https://proceedings.neurips.cc/paper_files/paper/2022/file/b1efde53be364a73914f58805a001731-Paper-Conference.pdf},
 volume = {35},
 year = {2022}
}

@inproceedings{DPO,
 author = {Rafailov, Rafael and Sharma, Archit and Mitchell, Eric and Manning, Christopher D and Ermon, Stefano and Finn, Chelsea},
 booktitle = {Advances in Neural Information Processing Systems},
 editor = {A. Oh and T. Naumann and A. Globerson and K. Saenko and M. Hardt and S. Levine},
 pages = {53728--53741},
 publisher = {Curran Associates, Inc.},
 title = {Direct Preference Optimization: Your Language Model is Secretly a Reward Model},
 url = {https://proceedings.neurips.cc/paper_files/paper/2023/file/a85b405ed65c6477a4fe8302b5e06ce7-Paper-Conference.pdf},
 volume = {36},
 year = {2023}
}

@article{Privacy-preserving_alignment_survey,
author = {Junwei Chen  and Pengpeng Qiao  and Wei Yang Bryan Lim  and Manjiang Yu  and Yang Cao },
title = {Privacy-Preserving LLM Preference Alignment: A Systematic Survey},
journal = {TechRxiv},
volume = {2026},
number = {0130},
pages = {},
year = {2026},
doi = {10.36227/techrxiv.176978434.43544529/v1},
URL = {https://www.techrxiv.org/doi/abs/10.36227/techrxiv.176978434.43544529/v1},
eprint = {https://www.techrxiv.org/doi/pdf/10.36227/techrxiv.176978434.43544529/v1}}

@inproceedings{KTO,
author = {Ethayarajh, Kawin and Xu, Winnie and Muennighoff, Niklas and Jurafsky, Dan and Kiela, Douwe},
title = {Model alignment as prospect theoretic optimization},
year = {2024},
publisher = {JMLR.org},
booktitle = {Proceedings of the 41st International Conference on Machine Learning},
articleno = {504},
numpages = {18},
location = {Vienna, Austria},
series = {ICML'24}
}

@misc{zhao2023slichfsequencelikelihoodcalibration,
      title={SLiC-HF: Sequence Likelihood Calibration with Human Feedback}, 
      author={Yao Zhao and Rishabh Joshi and Tianqi Liu and Misha Khalman and Mohammad Saleh and Peter J. Liu},
      year={2023},
      eprint={2305.10425},
      archivePrefix={arXiv},
      primaryClass={cs.CL},
      url={https://arxiv.org/abs/2305.10425}, 
}

@inproceedings{liu-etal-2025-lipo,
    title = "{L}i{PO}: Listwise Preference Optimization through Learning-to-Rank",
    author = "Liu, Tianqi  and
      Qin, Zhen  and
      Wu, Junru  and
      Shen, Jiaming  and
      Khalman, Misha  and
      Joshi, Rishabh  and
      Zhao, Yao  and
      Saleh, Mohammad  and
      Baumgartner, Simon  and
      Liu, Jialu  and
      Liu, Peter J  and
      Wang, Xuanhui",
    editor = "Chiruzzo, Luis  and
      Ritter, Alan  and
      Wang, Lu",
    booktitle = "Proceedings of the 2025 Conference of the Nations of the Americas Chapter of the Association for Computational Linguistics: Human Language Technologies (Volume 1: Long Papers)",
    month = apr,
    year = "2025",
    address = "Albuquerque, New Mexico",
    publisher = "Association for Computational Linguistics",
    url = "https://aclanthology.org/2025.naacl-long.121/",
    doi = "10.18653/v1/2025.naacl-long.121",
    pages = "2404--2420",
    ISBN = "979-8-89176-189-6"
}

@inproceedings{song2024pro,
author = {Song, Feifan and Yu, Bowen and Li, Minghao and Yu, Haiyang and Huang, Fei and Li, Yongbin and Wang, Houfeng},
title = {Preference ranking optimization for human alignment},
year = {2024},
isbn = {978-1-57735-887-9},
publisher = {AAAI Press},
url = {https://doi.org/10.1609/aaai.v38i17.29865},
doi = {10.1609/aaai.v38i17.29865},
booktitle = {Proceedings of the Thirty-Eighth AAAI Conference on Artificial Intelligence and Thirty-Sixth Conference on Innovative Applications of Artificial Intelligence and Fourteenth Symposium on Educational Advances in Artificial Intelligence},
articleno = {2117},
numpages = {9},
series = {AAAI'24/IAAI'24/EAAI'24}
}

@inproceedings{yuan2023rrhf,
 author = {Yuan, Hongyi and Yuan, Zheng and Tan, Chuanqi and Wang, Wei and Huang, Songfang and Huang, Fei},
 booktitle = {Advances in Neural Information Processing Systems},
 editor = {A. Oh and T. Naumann and A. Globerson and K. Saenko and M. Hardt and S. Levine},
 pages = {10935--10950},
 publisher = {Curran Associates, Inc.},
 title = {RRHF: Rank Responses to Align Language Models with Human Feedback},
 url = {https://proceedings.neurips.cc/paper_files/paper/2023/file/23e6f78bdec844a9f7b6c957de2aae91-Paper-Conference.pdf},
 volume = {36},
 year = {2023}
}

@inproceedings{zhao-etal-2025-permutative,
    title = "Permutative Preference Alignment from Listwise Ranking of Human Judgments",
    author = "Zhao, Yang  and
      Wang, Yixin  and
      Yin, Mingzhang",
    editor = "Christodoulopoulos, Christos  and
      Chakraborty, Tanmoy  and
      Rose, Carolyn  and
      Peng, Violet",
    booktitle = "Proceedings of the 2025 Conference on Empirical Methods in Natural Language Processing",
    month = nov,
    year = "2025",
    address = "Suzhou, China",
    publisher = "Association for Computational Linguistics",
    url = "https://aclanthology.org/2025.emnlp-main.17/",
    doi = "10.18653/v1/2025.emnlp-main.17",
    pages = "310--334",
    ISBN = "979-8-89176-332-6"
}

@article{yang2024qwen2,
  title={Qwen2 Technical Report},
  author={Yang, An and Yang, Baosong and Hui, Binyuan and Zheng, Bo and Yu, Bowen and Zhou, Chang and Li, Chengpeng and Li, Chengyuan and Liu, Dayiheng and Huang, Fei and others},
  journal={eprint arXiv: 2407.10671},
  year={2024}
}

@article{jiang2023mistral7b,
  title={Mistral 7B},
  author={Jiang, Albert Q and Sablayrolles, Alexandre and Mensch, Arthur and Bamford, Chris and Chaplot, Devendra Singh and Casas, Diego de las and Bressand, Florian and Lengyel, Gianna and Lample, Guillaume and Saulnier, Lucile and others},
  journal={arXiv preprint arXiv:2310.06825},
  year={2023}
}

@inproceedings{ding2023enhancing,
    title = "Enhancing Chat Language Models by Scaling High-quality Instructional Conversations",
    author = "Ding, Ning  and
      Chen, Yulin  andsong2024pro
      Xu, Bokai  and
      Qin, Yujia  and
      Hu, Shengding  and
      Liu, Zhiyuan  and
      Sun, Maosong  and
      Zhou, Bowen",
    editor = "Bouamor, Houda  and
      Pino, Juan  and
      Bali, Kalika",
    booktitle = "Proceedings of the 2023 Conference on Empirical Methods in Natural Language Processing",
    month = dec,
    year = "2023",
    address = "Singapore",
    publisher = "Association for Computational Linguistics",
    url = "https://aclanthology.org/2023.emnlp-main.183/",
    doi = "10.18653/v1/2023.emnlp-main.183",
    pages = "3029--3051"
}

@misc{li2023alpacaeval,
      title={Length-Controlled AlpacaEval: A Simple Way to Debias Automatic Evaluators}, 
      author={Yann Dubois and Balázs Galambosi and Percy Liang and Tatsunori B. Hashimoto},
      year={2025},
      eprint={2404.04475},
      archivePrefix={arXiv},
      primaryClass={cs.LG},
      url={https://arxiv.org/abs/2404.04475}, 
}

@inproceedings{zheng2023judging,
author = {Zheng, Lianmin and Chiang, Wei-Lin and Sheng, Ying and Zhuang, Siyuan and Wu, Zhanghao and Zhuang, Yonghao and Lin, Zi and Li, Zhuohan and Li, Dacheng and Xing, Eric P. and Zhang, Hao and Gonzalez, Joseph E. and Stoica, Ion},
title = {Judging LLM-as-a-judge with MT-bench and Chatbot Arena},
year = {2023},
publisher = {Curran Associates Inc.},
address = {Red Hook, NY, USA},
booktitle = {Proceedings of the 37th International Conference on Neural Information Processing Systems},
articleno = {2020},
numpages = {29},
location = {New Orleans, LA, USA},
series = {NIPS '23}
}

@article{achiam2023gpt,
  title={Gpt-4 technical report},
  author={Achiam, Josh and Adler, Steven and Agarwal, Sandhini and Ahmad, Lama and Akkaya, Ilge and Aleman, Florencia Leoni and Almeida, Diogo and Altenschmidt, Janko and Altman, Sam and Anadkat, Shyamal and others},
  journal={arXiv preprint arXiv:2303.08774},
  year={2023}
}

@InProceedings{yu2024dp,
  title = 	 {Privacy-Preserving Instructions for Aligning Large Language Models},
  author =       {Yu, Da and Kairouz, Peter and Oh, Sewoong and Xu, Zheng},
  booktitle = 	 {Proceedings of the 41st International Conference on Machine Learning},
  pages = 	 {57480--57506},
  year = 	 {2024},
  editor = 	 {Salakhutdinov, Ruslan and Kolter, Zico and Heller, Katherine and Weller, Adrian and Oliver, Nuria and Scarlett, Jonathan and Berkenkamp, Felix},
  volume = 	 {235},
  series = 	 {Proceedings of Machine Learning Research},
  month = 	 {21--27 Jul},
  publisher =    {PMLR},
  url = 	 {https://proceedings.mlr.press/v235/yu24e.html}
}

@article{wu2023privately,
  title={Privately aligning language models with reinforcement learning},
  author={Wu, Fan and Inan, Huseyin A and Backurs, Arturs and Chandrasekaran, Varun and Kulkarni, Janardhan and Sim, Robert},
  journal={arXiv preprint arXiv:2310.16960},
  year={2023}
}

@article{teku2025props,
  title={Props: Progressively private self-alignment of large language models},
  author={Teku, Noel and Tian, Fengwei and Bhattacharjee, Payel and Chakraborty, Souradip and Bedi, Amrit Singh and Tandon, Ravi},
  journal={arXiv preprint arXiv:2508.06783},
  year={2025}
}

@inproceedings{chowdhury2024differentially,
  title={Differentially private reward estimation with preference feedback},
  author={Chowdhury, Sayak Ray and Zhou, Xingyu and Natarajan, Nagarajan},
  booktitle={International Conference on Artificial Intelligence and Statistics},
  pages={4843--4851},
  year={2024},
  organization={PMLR}
}

@article{loshchilov2017decoupled,
  title={Decoupled weight decay regularization},
  author={Loshchilov, Ilya and Hutter, Frank},
  journal={arXiv preprint arXiv:1711.05101},
  year={2017}
}

@article{cui2023ultrafeedback,
  title={Ultrafeedback: Boosting language models with scaled ai feedback},
  author={Cui, Ganqu and Yuan, Lifan and Ding, Ning and Yao, Guanming and He, Bingxiang and Zhu, Wei and Ni, Yuan and Xie, Guotong and Xie, Ruobing and Lin, Yankai and others},
  journal={arXiv preprint arXiv:2310.01377},
  year={2023}
}

@book{chaudhuri2020randomized,
  title={Randomized response: Theory and techniques},
  author={Chaudhuri, Arijit and Mukerjee, Rahul},
  year={2020},
  publisher={Routledge}
}

@article{feng2024exposing,
  title={Exposing privacy gaps: Membership inference attack on preference data for LLM alignment},
  author={Feng, Qizhang and Kasa, Siva Rajesh and Kasa, Santhosh Kumar and Yun, Hyokun and Teo, Choon Hui and Bodapati, Sravan Babu},
  journal={arXiv preprint arXiv:2407.06443},
  year={2024}
}

@article{li2025listwise,
  title={Listwise Preference Alignment Optimization for Tail Item Recommendation},
  author={Li, Zihao and Yang, Chao and Zhang, Tong and Chen, Yakun and Wang, Xianzhi and Xu, Guandong and Dong, Daoyi},
  journal={IEEE Transactions on Computational Social Systems},
  year={2025},
  publisher={IEEE}
}

@inproceedings{wang-etal-2025-user,
    title = "User Feedback Alignment for {LLM}-powered Exploration in Large-scale Recommendation Systems",
    author = "Wang, Jianling  and
      Liu, Yifan  and
      Sun, Yinghao  and
      Ma, Xuejian  and
      Wang, Yueqi  and
      Ma, He  and
      Su, Zhengyang  and
      Chen, Minmin  and
      Gao, Mingyan  and
      Dalal, Onkar  and
      Chi, Ed H.  and
      Hong, Lichan  and
      Han, Ningren  and
      Lu, Haokai",
    editor = "Rehm, Georg  and
      Li, Yunyao",
    booktitle = "Proceedings of the 63rd Annual Meeting of the Association for Computational Linguistics (Volume 6: Industry Track)",
    month = jul,
    year = "2025",
    address = "Vienna, Austria",
    publisher = "Association for Computational Linguistics",
    url = "https://aclanthology.org/2025.acl-industry.70/",
    doi = "10.18653/v1/2025.acl-industry.70",
    pages = "996--1003",
    ISBN = "979-8-89176-288-6"
}

@inproceedings{lambdarank,
 author = {Burges, Christopher and Ragno, Robert and Le, Quoc},
 booktitle = {Advances in Neural Information Processing Systems},
 editor = {B. Sch\"{o}lkopf and J. Platt and T. Hoffman},
 pages = {},
 publisher = {MIT Press},
 title = {Learning to Rank with Nonsmooth Cost Functions},
 url = {https://proceedings.neurips.cc/paper_files/paper/2006/file/af44c4c56f385c43f2529f9b1b018f6a-Paper.pdf},
 volume = {19},
 year = {2006}
}

@article{NDCG,
author = {Valcarce, Daniel and Bellog\'{\i}n, Alejandro and Parapar, Javier and Castells, Pablo},
title = {Assessing ranking metrics in top-N recommendation},
year = {2020},
issue_date = {Aug 2020},
publisher = {Kluwer Academic Publishers},
address = {USA},
volume = {23},
number = {4},
issn = {1386-4564},
url = {https://doi.org/10.1007/s10791-020-09377-x},
doi = {10.1007/s10791-020-09377-x},
journal = {Inf. Retr.},
month = jun,
pages = {411–448},
numpages = {38}
}

@article{liu2025grpo,
  title={GRPO Privacy Is at Risk: A Membership Inference Attack Against Reinforcement Learning With Verifiable Rewards},
  author={Liu, Yule and Zhang, Heyi and Zheng, Jinyi and Sun, Zhen and Peng, Zifan and Cong, Tianshuo and Yang, Yilong and He, Xinlei and Ma, Zhuo},
  journal={arXiv preprint arXiv:2511.14045},
  year={2025}
}

@inproceedings{
    gao2026differentially,
    title={Differentially Private Preference Data Synthesis for Large Language Model Alignment},
    author={Fengyu Gao and Jing Yang},
    booktitle={Forty-third International Conference on Machine Learning},
    year={2026},
    url={https://openreview.net/forum?id=Ctk4XLaUpl}
}

@article{xu2025continuous,
    title={Continuous publication of weighted graphs with local differential privacy},
    author={Xu, Wen and Qiao, Pengpeng and Liu, Shang and Zheng, Zhirun and Cao, Yang and Li, Zhetao},
    journal={Proceedings of the VLDB Endowment},
    volume={18},
    number={11},
    pages={4214--4226},
    publisher={VLDB Endowment}
}

@article{li2026privacy,
    title={Privacy-Preserving Yet Vulnerable: Data Poisoning Attacks Against Differential Privacy Sparse Mobile Crowdsensing System},
    author={Li, Chengxin and Gu, Yujie and Qiao, Pengpeng and Pan, Shengli and Sakurai, Kouichi and Li, Zhetao},
    journal={IEEE Transactions on Mobile Computing},
    year={2026},
    publisher={IEEE}
}
}

\clearpage

%%%%%%%%%%%%%%%%%%%%%%%%%%%%%%%%%%%%%%%%%%%%%%%%%%%%%%%%%%%%

\appendix

\section{Additional Privacy Granularities}
\label{app:privacy-granularities}

This appendix records two stronger privacy granularities over listwise preference data. These definitions are provided for completeness only. The main analysis in the paper is based on item-level relocation adjacency in Definition~\ref{def:item-relocation-adjacency}.

\subsection{Rank-Level Local Differential Privacy}
\label{app:rank-level-ldp}

\begin{definition}[Rank-level replacement adjacency]
  \label{def:rank-level-adjacency}
  Two rankings \(\phi,\phi'\in S_m\) are rank-level adjacent, denoted by
  \[
    \phi \sim_{\mathrm{rank}} \phi',
  \]
  if they are two possible complete rankings over the same candidate set. Equivalently, \(\phi\) and \(\phi'\) may differ arbitrarily as permutations in \(S_m\).
\end{definition}

\begin{definition}[Rank-level local differential privacy on rankings]
  \label{def:rank-level-ranking-ldp}
  Let \(Q:S_m\to\mathcal{O}\) be a randomized mechanism over rankings. We say that \(Q\) satisfies \(\varepsilon\)-rank-level local differential privacy if for all rank-level adjacent rankings \(\phi,\phi'\in S_m\) and all outputs \(o\in\mathcal{O}\),
  \[
    \frac{Q(o\mid \phi)}{Q(o\mid \phi')}
    \le
    e^\varepsilon.
  \]
\end{definition}

Rank-level LDP protects the entire ordering for a single prompt. Unlike item-level LDP, which only requires indistinguishability under one item relocation, rank-level LDP requires indistinguishability between any two complete rankings over the same candidate set.

\subsection{User-Level Local Differential Privacy in Ranking}
\label{app:user-level-ldp}

For user-level privacy, the private input is not a single ranking but the user's full local preference profile. Let
\[
  \Phi_u = \bigl(\phi_{u,1},\ldots,\phi_{u,T_u}\bigr)
\]
denote the collection of listwise rankings provided by user \(u\), where \(\phi_{u,t}\in S_{m_t}\) is the ranking for the \(t\)-th prompt and candidate set.

\begin{definition}[User-level replacement adjacency]
  \label{def:user-level-adjacency}
  Two user preference profiles \(\Phi_u\) and \(\Phi'_u\) are user-level adjacent, denoted by
  \[
    \Phi_u \sim_{\mathrm{user}} \Phi'_u,
  \]
  if they are two possible complete preference profiles of the same user. Equivalently, \(\Phi_u\) and \(\Phi'_u\) may differ arbitrarily in any or all of their component rankings.
\end{definition}

\begin{definition}[User-level local differential privacy]
  \label{def:user-level-ldp}
  Let \(M:\mathcal{X}_{\mathrm{user}}\to\mathcal{O}\) be a randomized mechanism that takes a user's full preference profile as input. We say that \(M\) satisfies \(\varepsilon\)-user-level local differential privacy if for all user-level adjacent profiles \(\Phi_u,\Phi'_u\in\mathcal{X}_{\mathrm{user}}\) and all outputs \(o\in\mathcal{O}\),
  \[
    \frac{M(o\mid \Phi_u)}{M(o\mid \Phi'_u)}
    \le
    e^\varepsilon.
  \]
\end{definition}

User-level LDP protects the complete preference profile of a user, including all rankings submitted across prompts. This is stronger than protecting a single listwise ranking and generally requires privacy calibration at the profile level rather than at the single-ranking level.

\subsection{Relation to item-level LDP}
Definition~\ref{def:item-level-ranking-ldp} gives a per-list guarantee: for a
single prompt, two rankings that differ by relocating one item are
indistinguishable up to \(e^\varepsilon\). If the same user contributes
\(N\) listwise feedbacks and the mechanism is applied independently to each
list, sequential composition gives
\[
  \prod_{t=1}^N
  \frac{Q_t(o_t\mid \phi_{u,t})}{Q_t(o_t\mid \phi'_{u,t})}
  \le e^{N\varepsilon},
\]
whenever each pair \((\phi_{u,t},\phi'_{u,t})\) is item-level adjacent. Thus,
the per-list item-level guarantee induces a user-level privacy loss of
\(\varepsilon_{\mathrm{user}}=N\varepsilon\) for \(N\) released lists under this
adjacency. For example, \(N=10\) prompts with per-list \(\varepsilon=1\) gives
\(\varepsilon_{\mathrm{user}}=10\), which is already a loose worst-case bound.

For the stronger user-level replacement adjacency in
Definition~\ref{def:user-level-adjacency}, where each ranking may change
arbitrarily, the direct implication from item-level LDP is even more
conservative. Since any two rankings in \(S_m\) can be connected by at most
\(m-1\) item relocations, group privacy and sequential composition give the
worst-case bound
\[
  \varepsilon_{\mathrm{user}}
  \le
  N(m-1)\varepsilon
\]
for \(N\) lists of size \(m\). This illustrates that item-level LDP protects the main per-item position leakage considered in the paper, while direct extensions to rank-level or user-level privacy can incur large privacy-budget costs. We leave a tight user-level analysis as future work.

\section{Proofs}
\label{app:proofs}

\subsection{Sensitivity and Noise Scale Calculation}
\label{app:sensitivity-calculation}

To compare different release granularities on a unified scale, we report their \(L_1\) sensitivities and the corresponding Laplace noise scales. This analysis
is intended as a common-scale comparison rather than an optimal mechanism choice: ranking vectors and binary partitions may admit better discrete DP mechanisms,
so their Laplace noise scales serve as conservative upper bounds.

We provide the sensitivity and noise scale calculation for the three representation levels used in Section~\ref{subsec:representation-motivation}.
Throughout this section, sensitivity is measured in \(L_1\) distance:
\[
  \Delta_R
  =
  \sup_{x\sim x'}
  \|R(x)-R(x')\|_1,
\]
where \(x\sim x'\) denotes the item-level relocation adjacency in~\ref{subsec:item-level-ldp-ranking}.

\paragraph{Score representation.}

For the score representation, the released object is the cardinal score vector
\[
  R_0(\mathbf{s})=\mathbf{s}.
\]
We normalize the score range to \([0,m]\), so that
\[
  \mathbf{s}\in[0,m]^m.
\]
For two adjacent inputs \(\mathbf{s}\) and \(\mathbf{s}'\), the \(L_1\) change is
\[
  \|R_0(\mathbf{s})-R_0(\mathbf{s}')\|_1
  =
  \|\mathbf{s}-\mathbf{s}'\|_1
  =
  \sum_{i=1}^m |s_i-s_i'|.
\]
Since each coordinate lies in \([0,m]\), each coordinate can change by at most \(m\):
\[
  |s_i-s_i'|\le m.
\]
There are \(m\) coordinates, so
\[
  \sum_{i=1}^m |s_i-s_i'|
  \le
  \sum_{i=1}^m m
  =
  m^2.
\]
This bound is tight: for example, one may take
\[
  \mathbf{s}=(m,m,\ldots,m),
  \qquad
  \mathbf{s}'=(0,0,\ldots,0),
\]
which gives
\[
  \|\mathbf{s}-\mathbf{s}'\|_1=m^2.
\]
Therefore,
\[
  \Delta_{R_0}=m^2.
\]

\paragraph{Ranking representation.}

For the ranking representation, the released object is the full ranking encoded by fixed positional scores.
Let \(r_i(\phi)\in\{1,\ldots,m\}\) denote the rank position of item \(i\), where smaller rank means higher preference.
We encode the ranking as
\[
  \rho_i(\phi)=m-r_i(\phi),
\]
so that the top-ranked item has positional score \(m-1\), the bottom-ranked item has positional score \(0\), and adjacent rank positions differ by one.
Thus,
\[
  R_1(\phi)=\boldsymbol{\rho}(\phi)
  =
  (\rho_1(\phi),\ldots,\rho_m(\phi)).
\]

Under the local adjacency relation for rankings, two adjacent rankings differ by relocating one response from one rank position to another, while preserving the relative order of all other responses.
Suppose item \(i\) moves from rank position \(p\) to rank position \(q\).
Let
\[
  d=|p-q|
\]
be the number of positions it moves.

The moved item changes its positional score from
\[
  m-p
\]
to
\[
  m-q.
\]
Therefore, its absolute change is
\[
  |(m-p)-(m-q)|
  =
  |q-p|
  =
  d.
\]

The only other affected items are the \(d\) responses between positions \(p\) and \(q\).
Each of these responses shifts by exactly one rank position, and since adjacent positional scores differ by one, each contributes \(1\) to the \(L_1\) change.
Hence, the total contribution of the shifted intervening responses is
\[
  d.
\]

All remaining responses keep the same rank position and contribute \(0\) to the \(L_1\) change.
Therefore, the total \(L_1\) change is
\[
  d+d=2d.
\]
Since the largest possible relocation distance is
\[
  d_{\max}=m-1,
\]
we obtain
\[
  \|R_1(\phi)-R_1(\phi')\|_1
  \le
  2(m-1).
\]
This bound is tight.
For example, moving the top-ranked item to the bottom position gives \(d=m-1\), and the total \(L_1\) change is
\[
  2(m-1).
\]
Therefore,
\[
  \Delta_{R_1}=2(m-1).
\]

\paragraph{Binary partition representation.}

For the binary top-\(B\) representation, the released object is
\[
  R_2(\phi)=Y(\phi)\in\{0,1\}^m,
  \qquad
  Y_i(\phi)=\mathbf{1}[r_i(\phi)\le B].
\]
Thus, exactly \(B\) coordinates are equal to \(1\), and the remaining \(m-B\) coordinates are equal to \(0\).

Under the local adjacency relation for binary top-\(B\) labels, two adjacent labels differ by one top-\(B\) replacement: one response leaves the top-\(B\) group and one response enters it.
Equivalently, one coordinate changes from \(1\) to \(0\), and another coordinate changes from \(0\) to \(1\).
Therefore, the \(L_1\) change is
\[
  |1-0|+|0-1|=2.
\]
All other coordinates remain unchanged.
Hence,
\[
  \|R_2(\phi)-R_2(\phi')\|_1
  \le 2.
\]
The bound is tight whenever one item is swapped across the top-\(B\) boundary.
Therefore,
\[
  \Delta_{R_2}=2.
\]

Combining the three cases gives
\[
  \Delta_{R_0}=m^2,
  \qquad
  \Delta_{R_1}=2(m-1),
  \qquad
  \Delta_{R_2}=2.
\]
Consequently, under the Laplace mechanism, the coordinate-wise noise scales are
\[
  \frac{\Delta_{R_0}}{\varepsilon}
  =
  \frac{m^2}{\varepsilon},
  \qquad
  \frac{\Delta_{R_1}}{\varepsilon}
  =
  \frac{2(m-1)}{\varepsilon},
  \qquad
  \frac{\Delta_{R_2}}{\varepsilon}
  =
  \frac{2}{\varepsilon}.
\]

%------------------------------------------------------

\subsection{Sensitivity of the rank-aware utility}
\label{app:proof-rank-utility-sensitivity}

\begin{lemma}[Sensitivity of the linear rank-aware utility]
  \label{lem:rank-utility-sensitivity}
  Let
  \[
    U(\phi,z)
    =
    \sum_{i=1}^m z_i\left(B+\frac12-r_i(\phi)\right),
    \qquad
    z\in\mathcal Z_B,
  \]
  where \(\mathcal Z_B=\{z\in\{0,1\}^m:\sum_i z_i=B\}\). Under item-level
  relocation adjacency,
  \[
    \Delta
    :=
    \max_{\phi\sim\phi'}
    \max_{z\in\mathcal Z_B}
    |U(\phi,z)-U(\phi',z)|
    =
    \max\{B,m-B\}.
  \]
\end{lemma}

\begin{proof}
  Consider two adjacent rankings \(\phi\sim\phi'\), where \(\phi'\) is obtained
  from \(\phi\) by relocating one item \(x\) from rank \(a\) to rank \(b\). It
  suffices to consider the case \(a<b\), since the case \(a>b\) is symmetric.
  Let \(L=b-a\). Then item \(x\) moves down by \(L\) positions, while the \(L\)
  items between ranks \(a+1\) and \(b\) each move up by one position.

  For any fixed \(z\in\mathcal Z_B\), the constant term \(B+\frac12\) cancels,
  and therefore
  \[
    U(\phi,z)-U(\phi',z)
    =
    \sum_{i=1}^m z_i\bigl(r_i(\phi')-r_i(\phi)\bigr).
  \]
  Let \(I\) be the set of items shifted up by one position, and define
  \(c=\sum_{i\in I} z_i\). Then
  \[
    U(\phi,z)-U(\phi',z)
    =
    z_x L - c .
  \]

  If \(z_x=0\), then
  \[
    |U(\phi,z)-U(\phi',z)|=c\le \min\{B,L\}\le B.
  \]
  If \(z_x=1\), then \(B-1\) selected items other than \(x\) must be placed among
  the \(m-1\) remaining items. Since only \(m-L-1\) of these items lie outside
  the shifted interval \(I\), at least
  \[
    \max\{0,\,B-1-(m-L-1)\}
    =
    \max\{0,\,B+L-m\}
  \]
  selected items must lie in \(I\). Hence
  \[
    |U(\phi,z)-U(\phi',z)|
    =
    L-c
    \le
    L-\max\{0,\,B+L-m\}
    =
    \min\{L,m-B\}
    \le
    m-B.
  \]
  Combining the two cases gives
  \[
    |U(\phi,z)-U(\phi',z)|
    \le
    \max\{B,m-B\}.
  \]

  It remains to show that this bound is tight. Relocate the item at rank \(1\)
  to rank \(m\), so \(L=m-1\). If \(z_x=0\) and \(z\) selects \(B\) of the shifted
  items, then the absolute utility change is \(B\). If \(z_x=1\) and \(z\) selects
  \(x\) together with \(B-1\) shifted items, then the absolute utility change is
  \[
    (m-1)-(B-1)=m-B.
  \]
  Therefore the maximum possible change is exactly
  \[
    \Delta=\max\{B,m-B\}.
  \]
\end{proof}

\subsection{Privacy of the rank-aware release mechanism}
\label{app:rank-aware-release-privacy}

\begin{theorem}[Item-level privacy and tightness]
\label{thm:rank-aware-release-privacy}
Fix \(m\ge 2\), \(1\le B\le m-1\), and \(\eta\ge 0\).
Let
\[
  \mathcal{Z}_B
  =
  \left\{z\in\{0,1\}^{m}:\sum_{i=1}^{m}z_i=B\right\},
  \qquad
  U(\phi,z)
  =
  \sum_{i=1}^{m}z_i\left(B+\frac12-r_i(\phi)\right),
\]
and define
\[
  Q_\eta(z\mid\phi)
  =
  \frac{\exp\!\left(\eta U(\phi,z)\right)}{Z_\phi(\eta)},
  \qquad
  Z_\phi(\eta)
  =
  \sum_{z'\in\mathcal{Z}_B}\exp\!\left(\eta U(\phi,z')\right).
\]
Under item-level relocation adjacency,
\[
  \sup_{\phi\sim_{\mathrm{item}}\phi'}
  \max_{z\in\mathcal{Z}_B}
  \left|
    \log\frac{Q_\eta(z\mid\phi)}{Q_\eta(z\mid\phi')}
  \right|
  =
  \eta\max\{B,m-B\}.
\]
Consequently, choosing
\(\eta=\varepsilon/\max\{B,m-B\}\) yields
\(\varepsilon\)-item-level local differential privacy, and the
privacy bound is tight for \(\varepsilon>0\).
\end{theorem}

\begin{proof}
First, the normalizing constant is independent of the ranking.
Write \(a_r=B+\frac12-r\). For any fixed ranking \(\phi\), the map
\[
  z\longmapsto R_\phi(z)
  :=
  \{r_i(\phi):z_i=1\}
\]
is a bijection between \(\mathcal{Z}_B\) and the size-\(B\) subsets of
\(\{1,\ldots,m\}\). Therefore,
\[
  \begin{aligned}
    Z_\phi(\eta)
    &=
    \sum_{z\in\mathcal{Z}_B}
    \exp\!\left(\eta\sum_{i=1}^{m}z_i a_{r_i(\phi)}\right)\\
    &=
    \sum_{\substack{R\subseteq\{1,\ldots,m\}\\|R|=B}}
    \exp\!\left(\eta\sum_{r\in R}a_r\right)
    =: Z_{m,B}(\eta),
  \end{aligned}
\]
which depends only on \(m,B,\eta\), not on \(\phi\).

Next, consider adjacent rankings \(\phi,\phi'\).
By symmetry, it suffices to consider moving an item \(x\) from rank
\(a\) to rank \(b>a\). Put \(L=b-a\), and let \(I\) be the set of the
\(L\) items that move up by one rank. For a fixed output
\(z\in\mathcal{Z}_B\), set \(c=\sum_{i\in I}z_i\). Then
\[
  U(\phi,z)-U(\phi',z)=z_xL-c.
\]
If \(z_x=0\), the absolute difference is \(c\le B\).
If \(z_x=1\), it is \(L-c\), the number of unselected items in \(I\),
which is at most \(m-B\). Hence,
\[
  |U(\phi,z)-U(\phi',z)|\le \max\{B,m-B\}=:\Delta.
\]

Since all probabilities on \(\mathcal{Z}_B\) are positive and
\(Z_\phi(\eta)=Z_{\phi'}(\eta)\), we have
\[
  \begin{aligned}
    \left|
      \log\frac{Q_\eta(z\mid\phi)}{Q_\eta(z\mid\phi')}
    \right|
    &=
    \eta\,|U(\phi,z)-U(\phi',z)|\\
    &\le \eta\Delta.
  \end{aligned}
\]
Thus \(\eta=\varepsilon/\Delta\) implies
\(Q_\eta(z\mid\phi)\le e^\varepsilon Q_\eta(z\mid\phi')\)
for every adjacent pair and every output, establishing the privacy
guarantee.

To prove tightness, take
\[
  \phi=(1,2,\ldots,m),
  \qquad
  \phi'=(2,3,\ldots,m,1).
\]
These rankings are item-level adjacent. If \(z_1=0\) and \(z\)
selects \(B\) of the remaining items, then
\[
  U(\phi,z)-U(\phi',z)=-B.
\]
If \(z_1=1\) and \(z\) selects \(B-1\) of the remaining items, then
\[
  U(\phi,z)-U(\phi',z)=m-B.
\]
Both outputs are feasible because \(1\le B\le m-1\).
The largest absolute utility difference is therefore \(\Delta\),
so the largest absolute log-likelihood ratio is exactly
\(\eta\Delta\). This proves the claimed equality and tightness.
\end{proof}

\subsection{Derivation of the Flip Rate Formula}
\label{app:flip-rate-derivation}

For \(v\in\mathbb{R}^d\), define
\[
  \sigma_k(v):=\sum_{\substack{S\subseteq[d]\\ |S|=k}}\prod_{j\in S}v_j,
  \qquad
  \sigma_0(v)=1,
  \qquad
  \sigma_k(v)=0\ \text{if } k<0 \text{ or } k>d.
\]
Under the rank-aware exponential mechanism,
\[
  \Pr(\widetilde z=z\mid\phi)
  =
  \frac{\exp(\eta U(\phi,z))}
  {\sum_{z'\in\mathcal Z_B}\exp(\eta U(\phi,z'))},
  \qquad
  U(\phi,z)=\sum_{j=1}^m z_j w_j(\phi).
\]
Since
\[
  \exp(\eta U(\phi,z))=\prod_{j:z_j=1}e^{\eta w_j(\phi)},
  \qquad
  \sum_{z'\in\mathcal Z_B}\exp(\eta U(\phi,z'))
  =
  \sigma_B(e^{\eta w(\phi)}),
\]
the marginal probability that \(y_i\) is released as positive is
\[
  q_i(\phi):=\Pr(\widetilde z_i=1\mid\phi)
  =
  \frac{
    e^{\eta w_i(\phi)}
    \sigma_{B-1}(e^{\eta w_{-i}(\phi)})
  }{
    \sigma_B(e^{\eta w(\phi)})
  }.
\]

Using
\[
  \sigma_B(e^{\eta w(\phi)})
  =
  e^{\eta w_i(\phi)}\sigma_{B-1}(e^{\eta w_{-i}(\phi)})
  +
  \sigma_B(e^{\eta w_{-i}(\phi)}),
\]
we obtain
\[
  \frac{q_i(\phi)}{1-q_i(\phi)}
  =
  e^{\eta w_i(\phi)}
  \frac{\sigma_{B-1}(e^{\eta w_{-i}(\phi)})}
  {\sigma_B(e^{\eta w_{-i}(\phi)})}.
\]
Holding \(w_{-i}(\phi)\) fixed, this odds ratio is increasing in
\(w_i(\phi)\). Thus \(q_i(\phi)\) is monotone increasing in the rank-position
weight.

Finally, with \(z_i(\phi)=\mathbf{1}[r_i(\phi)\le B]\), the item-wise flip rate is
\[
  \rho_i(\phi)
  :=
  \Pr(\widetilde z_i\ne z_i(\phi)\mid\phi)
  =
  \begin{cases}
    1-q_i(\phi), & r_i(\phi)\le B,\\[1mm]
    q_i(\phi), & r_i(\phi)>B.
  \end{cases}
\]

\subsection{Proof of Theorem~\ref{thm:linear-extreme-reliability}}
\label{app:proof-optimal-score-function}

We prove that for any monotone rank score \(w_1\ge\cdots\ge w_m\),
\[
  \frac{S_w}{\Delta_w}\le \frac{m-1}{\max\{B,m-B\}},
\]
with equality iff \(w_s=\alpha-\beta s\) for some \(\beta>0\).

\begin{proof}
  Let
  \[
    \delta_s:=w_s-w_{s+1},\quad s\in[m-1],
    \qquad
    \delta_{(1)}\ge\cdots\ge\delta_{(m-1)}
  \]
  be the adjacent score gaps in decreasing order. Consider relocating one item
  from rank \(a\) to rank \(b>a\); the reverse case is symmetric. Let \(i_s\) be
  the item originally at rank \(s\). Then \(i_a\) is moved, while
  \(i_{a+1},\ldots,i_b\) shift forward by one rank. For any \(z\in\mathcal Z_B\),
  \[
    U_w(\phi,z)-U_w(\phi',z)
    =
    \sum_{\ell=a}^{b-1}
    \delta_\ell\bigl(z_{i_a}-z_{i_{\ell+1}}\bigr).
  \]

  If \(z_{i_a}=0\), at most \(B\) selected shifted items contribute. If
  \(z_{i_a}=1\), at most \(m-B\) unselected shifted items contribute. Hence
  \[
    |U_w(\phi,z)-U_w(\phi',z)|
    \le
    \max\left\{
      \sum_{j=1}^{B}\delta_{(j)},
      \sum_{j=1}^{m-B}\delta_{(j)}
    \right\}.
  \]
  Let \(K:=\max\{B,m-B\}\). The bound is tight by relocating an item across the
  full ranking and choosing \(z\) so that the contributing shifted items correspond
  to the largest \(K\) gaps. Therefore,
  \[
    \Delta_w=\sum_{j=1}^{K}\delta_{(j)}.
  \]

  Since
  \[
    w_1-w_m=\sum_{s=1}^{m-1}\delta_s
  \]
  and the average of the largest \(K\) gaps is at least the average of all gaps,
  \[
    \Delta_w
    =
    \sum_{j=1}^{K}\delta_{(j)}
    \ge
    \frac{K}{m-1}\sum_{s=1}^{m-1}\delta_s
    =
    \frac{K}{m-1}(w_1-w_m).
  \]
  Rearranging gives
  \[
    \frac{w_1-w_m}{\Delta_w}
    \le
    \frac{m-1}{K}
    =
    \frac{m-1}{\max\{B,m-B\}}.
  \]

  Equality holds iff all adjacent gaps are equal,
  \[
    \delta_1=\cdots=\delta_{m-1},
  \]
  which is equivalent to \(w_s=\alpha-\beta s\). Strict monotonicity gives
  \(\beta>0\).
\end{proof}

\subsection{Calibration of Randomized Response in Listwise Preference Data}

\label{app:rr-calibration}

Recall the clean top-\(B\) label vector
\[
  z_i(\phi)=\mathbf{1}[r_i(\phi)\le B],
  \qquad
  z(\phi)=(z_1(\phi),\ldots,z_m(\phi)).
\]
RR with per-coordinate privacy parameter \(\varepsilon_0\)
releases \(\tilde z_{\mathsf{RR}}\in\{0,1\}^m\) independently across
coordinates:
\[
  \Pr[(\tilde z_{\mathsf{RR}})_i=z_i(\phi)\mid\phi]
  =
  \frac{e^{\varepsilon_0}}{1+e^{\varepsilon_0}},
  \qquad
  \Pr[(\tilde z_{\mathsf{RR}})_i\ne z_i(\phi)\mid\phi]
  =
  \frac{1}{1+e^{\varepsilon_0}}.
\]

Under item-level relocation adjacency, the clean label vector changes in at most
two coordinates:
\[
  \Delta_H
  :=
  \sup_{\phi\sim_{\mathrm{item}}\phi'}
  \|z(\phi)-z(\phi')\|_0
  =
  2.
\]
Indeed, if one item enters the top-\(B\) set, one item must leave it, and vice
versa.

For any \(u\in\{0,1\}^m\), the RR likelihood ratio factorizes as
\[
  \frac{\Pr[\tilde z_{\mathsf{RR}}=u\mid\phi]}
  {\Pr[\tilde z_{\mathsf{RR}}=u\mid\phi']}
  =
  \prod_{i=1}^m
  \frac{
    \Pr[(\tilde z_{\mathsf{RR}})_i=u_i\mid z_i(\phi)]
  }{
    \Pr[(\tilde z_{\mathsf{RR}})_i=u_i\mid z_i(\phi')]
  }.
\]
Unchanged coordinates contribute ratio \(1\), and each changed coordinate
contributes at most \(e^{\varepsilon_0}\). Since at most two coordinates change,
\[
  \frac{\Pr[\tilde z_{\mathsf{RR}}=u\mid\phi]}
  {\Pr[\tilde z_{\mathsf{RR}}=u\mid\phi']}
  \le
  e^{2\varepsilon_0}.
\]
Choosing \(\varepsilon_0=\varepsilon/2\) yields
\[
  \frac{\Pr[\tilde z_{\mathsf{RR}}=u\mid\phi]}
  {\Pr[\tilde z_{\mathsf{RR}}=u\mid\phi']}
  \le
  e^{\varepsilon},
\]
so RR is \(\varepsilon\)-LDP under item-level relocation adjacency.

Substituting \(\varepsilon_0=\varepsilon/2\), the RR flip rate is
\[
  \rho_i^{\mathsf{RR}}(\phi)
  :=
  \Pr[(\tilde z_{\mathsf{RR}})_i\ne z_i(\phi)\mid\phi]
  =
  \frac{1}{1+e^{\varepsilon/2}},
  \qquad
  i\in[m].
\]

\section{Experiments Details}
\label{app:experiments-details}

\subsection{Training Details}
\label{app:training-details}

\noindent
The training hyperparameters for RR+KTO and our method on Qwen2-0.5B and Mistral-7B are shown in~\Cref{tab:training_details}. For non-KTO baselines, PPA and SLiC, we use the best hyperparameters reported in their original papers. Using 4 NVIDIA H100 80GB GPUs, each run takes about 2 hours for Qwen2-0.5B and about 8 hours for Mistral-7B.

\begin{table}[H]
  \centering
  \caption{Training details for private KTO experiments.}
  \label{tab:training_details}
  \footnotesize
  \setlength{\tabcolsep}{18pt}
  \renewcommand{\arraystretch}{1.1}
  \begin{tabular}{lc}
    \toprule
    \textbf{Hyperparameter} & \textbf{Value} \\
    \midrule
    Training dataset & ListUltraFeedback \\
    Number of candidates $m$ & 8 \\
    Top boundary $B$ & 4 \\
    Privacy budgets $\varepsilon$ & 0.5, 1, 3 \\
    KTO $\beta$ & 0.1 \\
    KTO $\lambda_D, \lambda_U$ & 1.0, 1.0 \\
    Mini batch / GPU & 2 \\
    Gradient accumulation steps & 8 \\
    GPUs & 4 \\
    Effective batch size & 64 \\
    Learning rate & $5\times10^{-6}$ \\
    Epochs & 1 \\
    Max prompt length & 512 \\
    Max response length & 1024 \\
    Optimizer & AdamW \\
    LR scheduler & Cosine \\
    Warmup ratio & 0.1 \\
    Weight decay & 0.01 \\
    Max grad norm & 1.0 \\
    DeepSpeed & ZeRO-2 \\
    Random seed & 42 \\
    \bottomrule
  \end{tabular}
\end{table}

\subsection{Evaluation Details}
\label{app:evaluation-details}

Table~\ref{tab:datasets_and_judges} reports the datasets and judge models used in our experiments. UltraChat200k is used for SFT, ListUltraFeedback$_{\rm train}$ for preference alignment, and ListUltraFeedback$_{\rm test}$ for in-domain proxy evaluation with ArmoRM and the RLHFlow pair-preference model. AlpacaEval and
MT-Bench provide general instruction-following benchmarks judged by GPT-4 Turbo.

\begin{table}[H]
  \centering
  \caption{Datasets and judge models used in our experiments.}
  \label{tab:datasets_and_judges}
  \footnotesize
  \setlength{\tabcolsep}{12pt}
  \renewcommand{\arraystretch}{1.05}
  \begin{tabular}{@{}lcc@{}}
    \toprule
    \textbf{Datasets} & \textbf{Examples} & \textbf{Judge Model} \\
    \midrule
    \textbf{UltraChat200k} & 207,865 & -- \\
    \textbf{ListUltraFeedback}$_{\text{train}}$ & 59,880 & -- \\
    \midrule
    \textbf{ListUltraFeedback}$_{\text{test}}$ & 1,968 & RLHFlow Pair-Preference, ArmoRM \\
    \textbf{AlpacaEval} & 805 & GPT-4 Turbo \\
    \textbf{MT-Bench} & 80 & GPT-4 Turbo \\
    \bottomrule
  \end{tabular}
\end{table}

\section{Additional Experiments}
\label{app:additional-experiments}

We report additional experiments on alternative alignment objectives,
variation across training seeds, and controlled mechanism ablations.
The four alignment metrics
follow the evaluation protocol in the main text. All scores are
reported as percentages, with higher values indicating better
performance. For the alignment evaluations, Avg.\ denotes the
unweighted mean of Pair-Preference, Scoring, AlpacaEval, and MT-Bench.

\subsection{Compatibility with pairwise alignment}
\label{app:additional-dpo}

RAEM releases a privatized top-\(B\) partition independently of the
downstream alignment objective. To apply DPO, we construct preference
pairs using only the released positive and negative groups: the
preferred response is drawn from the released positive group and the
rejected response from the released negative group. The admissible
pair set is
\[
  \mathcal{P}(\tilde z)
  =
  \left\{(y_i,y_j):\tilde z_i=1,\ \tilde z_j=0\right\}.
\]
Because all pairs are constructed only from the released partition
and the candidate responses, this conversion is post-processing and
incurs no additional privacy cost. The same construction is also
applicable to pairwise supervision for reward-model training in RLHF.

Table~\ref{tab:additional-dpo} compares RAEM followed by KTO or DPO on
Qwen2-0.5B at \(\varepsilon\in\{1,3\}\). RAEM is compatible with both
objectives, while RAEM+KTO achieves the higher average score at both
privacy budgets.

\begin{table}[t]
  \centering
  \small
  \caption{RAEM with KTO and DPO on Qwen2-0.5B. Bold values indicate
  the higher average score at each privacy budget.}
  \label{tab:additional-dpo}
  \setlength{\tabcolsep}{5pt}
  \resizebox{\linewidth}{!}{%
  \begin{tabular}{lrrrrrr}
    \toprule
    Method & \(\varepsilon\) & Pair-Preference & Scoring
           & AlpacaEval & MT-Bench & Avg. \\
    \midrule
    RAEM+KTO (ours) & 1 & 61.2 & 49.1 & 50.6 & 50.6 & \textbf{52.9} \\
    RAEM+DPO        & 1 & 58.1 & 48.5 & 48.9 & 53.5 & 52.3 \\
    \midrule
    RAEM+KTO (ours) & 3 & 66.4 & 54.6 & 57.8 & 59.7 & \textbf{59.6} \\
    RAEM+DPO        & 3 & 64.8 & 51.1 & 54.2 & 54.5 & 56.1 \\
    \bottomrule
  \end{tabular}%
  }
\end{table}

We retain KTO as the primary alignment objective for two reasons.
First, its unary supervision format directly matches the released
top-\(B\) partition, whereas DPO requires conversion into pairwise
comparisons. Second, the sigmoid utility in KTO attenuates the gradient
weight of examples with extreme policy--reference log-ratios relative
to the detached KL baseline. This includes strongly contradictory
examples, such as an undesirable response with a very high implied
reward or a desirable response with a very low implied reward.
Downweighting such examples provides a motivation for using KTO with
privacy-corrupted labels, and the results in
Table~\ref{tab:additional-dpo} support this choice in the evaluated
setting.

\subsection{Variation across training seeds}
\label{app:additional-seeds}

We conduct additional experiments on Mistral-7B at
\(\varepsilon=1\) using training seeds 123 and 456, in addition to seed
42 used in the main results. Table~\ref{tab:additional-seeds} reports
the complete results for RR+KTO and RAEM+KTO. The summary rows show the
mean and sample standard deviation over the three training seeds.
For Avg., we first compute the average of the four metrics for each
seed and then compute its mean and sample standard deviation across
seeds.

\begin{table}[t]
  \centering
  \small
  \caption{Mistral-7B results at \(\varepsilon=1\) across three
  training seeds. Summary rows report mean \(\pm\) sample standard
  deviation (\(n=3\)). Bold per-seed averages identify the better
  method for each seed; bold summary means identify the higher mean
  in each metric.}
  \label{tab:additional-seeds}
  \setlength{\tabcolsep}{4pt}
  \resizebox{\linewidth}{!}{%
  \begin{tabular}{llrrrrr}
    \toprule
    Seed & Method & Pair-Preference & Scoring
         & AlpacaEval & MT-Bench & Avg. \\
    \midrule
    42 (original) & RR+KTO   & 89.70 & 90.90 & 86.40 & 71.80 & 84.70 \\
    42 (original) & RAEM+KTO & 90.50 & 91.00 & 87.50 & 72.10 & \textbf{85.28} \\
    \midrule
    123 & RR+KTO   & 90.21 & 91.02 & 86.48 & 70.20 & 84.48 \\
    123 & RAEM+KTO & 91.11 & 91.22 & 88.05 & 70.44 & \textbf{85.21} \\
    \midrule
    456 & RR+KTO   & 90.77 & 90.15 & 86.17 & 70.89 & 84.50 \\
    456 & RAEM+KTO & 91.08 & 90.94 & 87.02 & 73.58 & \textbf{85.66} \\
    \midrule
    -- & RR+KTO, mean
       & \(90.23\pm0.54\)
       & \(90.69\pm0.47\)
       & \(86.35\pm0.16\)
       & \(70.96\pm0.80\)
       & \(84.56\pm0.12\) \\
    -- & RAEM+KTO, mean
       & \(\mathbf{90.90}\pm0.34\)
       & \(\mathbf{91.05}\pm0.15\)
       & \(\mathbf{87.52}\pm0.52\)
       & \(\mathbf{72.04}\pm1.57\)
       & \(\mathbf{85.38}\pm0.24\) \\
    \bottomrule
  \end{tabular}%
  }
\end{table}

RAEM+KTO achieves a higher average score than RR+KTO for each of the
three seeds. Its mean average score is \(85.38\), compared with
\(84.56\) for RR+KTO, an improvement of \(0.82\) percentage points.
These runs support the consistency of the observed advantage under
training-seed variation for Mistral-7B at this privacy budget.

\subsection{Controlled mechanism ablations}
\label{app:additional-ablations}

We compare two additional ablations on Qwen2-0.5B to examine the roles
of fixed cardinality, joint sampling, and within-group rank
information. All methods use the same training and evaluation setup.

\begin{enumerate}
  \item \textbf{RR + uniform fixed-\(B\) + KTO.}
  We apply coordinate-wise RR, followed by randomized post-processing
  that enforces exactly \(B\) positive labels. This step uses only the
  RR output and randomness independent of the clean ranking, and
  therefore incurs no additional privacy cost.

  \item \textbf{Top-membership EM + KTO.}
  We use an exponential mechanism on the same fixed-cardinality
  output space \(\mathcal{Z}_B\) as RAEM, with a utility that depends
  only on clean top-\(B\) membership. All items within the same clean
  group receive the same weight, so the utility does not use
  within-group ranks.
\end{enumerate}

Table~\ref{tab:additional-ablations} reports the results at
\(\varepsilon=0.5\) and \(\varepsilon=1\), together with the
coordinate-wise RR baseline and our linear rank-aware mechanism.

\begin{table}[t]
  \centering
  \small
  \caption{Controlled mechanism ablations on Qwen2-0.5B.
  All methods use KTO for downstream alignment. Bold values identify
  the highest average score at each privacy budget.}
  \label{tab:additional-ablations}
  \setlength{\tabcolsep}{5pt}
  \resizebox{\linewidth}{!}{%
  \begin{tabular}{lrrrrr}
    \toprule
    Method & Pair-Preference & Scoring & AlpacaEval & MT-Bench & Avg. \\
    \midrule
    \multicolumn{6}{l}{\(\varepsilon=0.5\)} \\
    \midrule
    Coordinate-wise RR + KTO
      & 52.9 & 38.6 & 48.0 & 52.5 & 48.0 \\
    RR + uniform fixed-\(B\) + KTO
      & 53.1 & 39.1 & 48.1 & 49.8 & 47.5 \\
    Top-membership EM + KTO
      & 52.9 & 40.1 & 49.5 & 48.8 & 47.8 \\
    Linear rank-aware EM + KTO (ours)
      & 59.2 & 43.6 & 51.8 & 49.7 & \textbf{51.1} \\
    \midrule
    \multicolumn{6}{l}{\(\varepsilon=1\)} \\
    \midrule
    Coordinate-wise RR + KTO
      & 57.9 & 43.7 & 50.7 & 52.0 & 51.1 \\
    RR + uniform fixed-\(B\) + KTO
      & 56.9 & 43.0 & 50.8 & 50.5 & 50.3 \\
    Top-membership EM + KTO
      & 58.1 & 45.1 & 50.1 & 50.2 & 50.9 \\
    Linear rank-aware EM + KTO (ours)
      & 61.2 & 49.1 & 50.6 & 50.6 & \textbf{52.9} \\
    \bottomrule
  \end{tabular}%
  }
\end{table}

Fixed-cardinality post-processing alone does not improve the average
score over coordinate-wise RR at either budget. The top-membership
exponential mechanism also remains below the linear rank-aware
mechanism, despite using the same fixed-cardinality output space and
joint exponential-mechanism sampling. Based on the displayed average
scores, the linear rank-aware mechanism improves over top-membership
EM by \(3.3\) percentage points at \(\varepsilon=0.5\) and \(2.0\)
percentage points at \(\varepsilon=1\).

These controlled comparisons support the benefit of using
within-group rank information in the evaluated setting. The results
are consistent with the design of RAEM: concentrating noise near the
top-\(B\) boundary helps preserve higher-quality preference signals
for downstream alignment.

\section{Supplementary Results}
\label{app:supplementary-results}

\begin{table}[h]
  \centering
  \caption{Qualitative Qwen2-0.5B examples from AlpacaEval where our method produces a more correct, relevant, or complete response than RR. Scores in parentheses are AlpacaEval GPT-4 Turbo preference scores against the same SFT reference; higher is better. Responses are summarized for space.}
  \label{tab:alpaca_eval_qwen_ours_beats_rr_samples}
  \large
  \setlength{\tabcolsep}{8pt}
  \renewcommand{\arraystretch}{1.22}
  \begin{tabularx}{\textwidth}{@{}>{\raggedright\arraybackslash}p{0.13\textwidth}>{\raggedright\arraybackslash}X@{}}
    \toprule

    % AlpacaEval example_id: 485
    \textbf{Prompt} &
    Given an algorithm or a coding search query, provide a related code snippet.
    Search query: \texttt{Python loop}. \\
    \midrule
    \textbf{$\varepsilon=1$} & \\
    RR &
    \textit{(1.004)}
    Gives \texttt{for I in range(10): print(i)}, which raises a
    \texttt{NameError} because Python identifiers are case-sensitive, and then
    incorrectly describes the query as ``10.'' \\
    Ours &
    \textit{(2.000)}
    Provides a valid \texttt{search\_list} function that iterates over a list,
    returns the index of the queried element when found, and returns $-1$
    otherwise. \\

    \midrule

    % AlpacaEval example_id: 528
    \textbf{Prompt} &
    Write a thematic tweet announcing that applications for summer internships
    are open. Start with a Halloween tie-in. \\
    \midrule
    \textbf{$\varepsilon=1$} & \\
    RR &
    \textit{(1.004)}
    Produces a generic internship announcement with a celebration emoji, but
    the tweet itself contains no Halloween-related opening or theme. \\
    Ours &
    \textit{(2.000)}
    Opens with ``Halloween is just around the corner!'' and then clearly
    announces that summer internship applications are open, satisfying both
    requested elements. \\

    \midrule

    % AlpacaEval example_id: 405
    \textbf{Prompt} &
    Write a JavaScript function that takes an array of five numbers, checks
    whether any element is prime, and prints every prime number. \\
    \midrule
    \textbf{$\varepsilon=3$} & \\
    RR &
    \textit{(1.002)}
    Prints an element only when both it and the following element are prime,
    and iterates only to \texttt{arr.length - 1}. It therefore misses valid
    primes, including a prime in the final position. \\
    Ours &
    \textit{(2.000)}
    Implements a correct primality test, examines every array element
    independently, and prints each element for which the test returns true. \\

    \midrule

    % AlpacaEval example_id: 293
    \textbf{Prompt} &
    Write a response that disagrees with the post:
    ``Technology is everything that doesn't work yet.'' \\
    \midrule
    \textbf{$\varepsilon=3$} & \\
    RR &
    \textit{(1.024)}
    Misreads the task as a request to interpret or rewrite the quotation and
    does not present a clear counterargument to it. \\
    Ours &
    \textit{(2.000)}
    Directly disagrees and supports the response with concrete examples of
    technologies already working in practice, including AI applications,
    e-commerce, and renewable energy. \\

    \bottomrule
  \end{tabularx}
\end{table}

\begin{table}[h]
  \centering
  \caption{Qualitative Mistral-7B examples from AlpacaEval where our method produces a more correct, relevant, or complete response than RR. Scores in parentheses are AlpacaEval GPT-4 Turbo preference scores against the same SFT reference; higher is better. Responses are summarized for space.}
  \label{tab:alpaca_eval_mistral_ours_beats_rr_samples}
  \large
  \setlength{\tabcolsep}{14pt}
  \renewcommand{\arraystretch}{1.42}
  \begin{tabularx}{\textwidth}{@{}>{\raggedright\arraybackslash}p{0.13\textwidth}>{\raggedright\arraybackslash}X@{}}
    \toprule

    % AlpacaEval example_id: 118
    \textbf{Prompt} &
    Does the dinosaurs really exist? \\
    \midrule
    \textbf{$\varepsilon=0.5$} & \\
    RR &
    \textit{(1.000)}
    Initially frames dinosaur existence as a matter of belief and
    interpretation, weakening an otherwise scientific answer. \\
    Ours &
    \textit{(2.000)}
    States the scientific consensus directly and supports it with fossilized
    bones, footprints, anatomical reconstruction, and examples of documented
    dinosaur species. \\

    \midrule

    % AlpacaEval example_id: 688
    \textbf{Prompt} &
    Suggest product bundles that are often purchased with a MacBook Air. \\
    \midrule
    \textbf{$\varepsilon=0.5$} & \\
    RR &
    \textit{(1.097)}
    Recommends an Apple Pencil on the mistaken premise that the MacBook Air has
    a touchscreen and describes several adapters inaccurately as cables. \\
    Ours &
    \textit{(2.000)}
    Suggests compatible and practically useful accessories, including
    AppleCare+, an external drive, a USB-C hub, headphones, a mouse, and a
    protective sleeve or backpack. \\

    \midrule

    % AlpacaEval example_id: 750
    \textbf{Prompt} &
    As a sports commentator, describe the winning play in the final seconds of
    a championship game. \\
    \midrule
    \textbf{$\varepsilon=0.5$} & \\
    RR &
    \textit{(1.024)}
    Begins a football scenario but is repetitive and terminates in the middle
    of the decisive play, leaving the requested winning moment unresolved. \\
    Ours &
    \textit{(2.000)}
    Delivers a complete and coherent basketball sequence: a tie game, ten
    seconds remaining, the final possession, a made three-pointer, and the
    resulting championship celebration. \\

    \bottomrule
  \end{tabularx}
\end{table}

% \clearpage

% \newpage
% \input{checklist.tex}

\end{document}